%% file: example_paper.tex
\documentclass{article}

\usepackage{microtype}
\usepackage{graphicx}
\usepackage{subcaption}
\usepackage{booktabs} 

\usepackage{hyperref}

\usepackage[preprint]{icml2026}

\usepackage{algorithm}
\usepackage{algpseudocode}
\usepackage{amsmath}
\usepackage{amssymb}
\usepackage{mathtools}
\usepackage{amsthm}

\usepackage{url}
\usepackage{graphicx}
\usepackage{caption}
\usepackage{booktabs}
\usepackage{multirow}
\usepackage{amssymb, amsfonts, bm, microtype}
\usepackage{array}
\usepackage{wrapfig}
\usepackage{tabularx}
\usepackage[table]{xcolor}

\usepackage[capitalize,noabbrev]{cleveref}

\definecolor{darkpink}{RGB}{255, 20, 147}
\definecolor{rho_EOS_red}{HTML}{B22222}
\definecolor{daedal_blue}{HTML}{29829F}
\definecolor{baseline_orange}{HTML}{E48246}
\definecolor{lightgreen}{HTML}{ECF6E4}
\definecolor{lightbluegreen}{HTML}{F0F9F8}
\definecolor{lightblue}{HTML}{E0F2FE}
\definecolor{lightred}{HTML}{FEE0E0}
\definecolor{lightorange}{HTML}{FEEBCB}
\definecolor{bleudefrance}{HTML}{5D90F4}

\theoremstyle{plain}

\theoremstyle{definition}

\theoremstyle{remark}

\usepackage[textsize=tiny]{todonotes}

\icmltitlerunning{Submission and Formatting Instructions for ICML 2026}

\begin{document}

\twocolumn[
  \icmltitle{$\rho$-\texttt{EOS}: Training-free Bidirectional Variable-Length Control \\ for Masked Diffusion LLMs}



  \icmlsetsymbol{equal}{*}
  \icmlsetsymbol{corr}{$\dagger$}

  \begin{icmlauthorlist}
    \icmlauthor{Jingyi Yang}{shi-ai-lab,fdu}
    \icmlauthor{Yuxian Jiang}{shi-ai-lab,fdu}
    \icmlauthor{Jing Shao}{shi-ai-lab,corr}
  \end{icmlauthorlist}

  \icmlaffiliation{shi-ai-lab}{Shanghai Artificial Intelligence Laboratory}
  \icmlaffiliation{fdu}{Fudan University}

  \icmlcorrespondingauthor{Jing Shao}{shaojing@pjlab.org.cn}

  \icmlkeywords{Machine Learning, ICML}

  \vskip 0.3in
]



\printAffiliationsAndNotice{}  

\begin{abstract}
Beyond parallel generation and global context modeling, current masked diffusion large language models (masked dLLMs, i.e., LLaDA) suffer from a fundamental limitation: they require a predefined, fixed generation length, which lacks flexibility and forces an inevitable trade-off between output quality and computational efficiency.
To address this, we study the denoising dynamics and find that the implicit density ($\rho$) of end-of-sequence (\texttt{EOS}) tokens serves as a reliable signal of generation sufficiency. 
In particular, the evolving implicit \texttt{EOS} density during denoising reveals whether the current masked space is excessive or insufficient, thereby guiding the adjustment direction for generation length.
Building on this insight, we propose \textbf{$\rho$-\texttt{EOS}}, a training-free, single-stage strategy that enables bidirectional variable-length generation for masked dLLMs.
Unlike prior two-stage approaches--which require separate length adjustment and iterative mask insertion phases while supporting only unidirectional expansion--\textbf{$\rho$-\texttt{EOS}} achieves bidirectional length adjustment within a unified denoising process by continuously estimating the implicit \texttt{EOS} density: excessively high density triggers \texttt{MASK} token contraction, while insufficient density induces expansion.
Extensive experiments on mathematics and code benchmarks demonstrate that \textbf{$\rho$-\texttt{EOS}} achieves comparable performance while substantially improving inference efficiency and token utilization. Code is available at \href{https://github.com/yjyddq/rho-EOS}{\textcolor{darkpink}{https://github.com/yjyddq/rho-EOS}}.
\end{abstract}

\input{sections/introduction}

\input{sections/related_work}

\input{sections/methods}

\input{sections/experiments}
\input{sections/conclusion}




\section*{Impact Statement}


This paper presents work whose goal is to advance the field of Machine
Learning. There are many potential societal consequences of our work, none
which we feel must be specifically highlighted here.


\nocite{langley00}

\bibliography{example_paper}
\bibliographystyle{icml2026}

\newpage
\appendix
\onecolumn
\section{Appendix}



\subsection{\textbf{Why does $\rho$-\texttt{EOS} work? -- Implicit length awareness in masked dLLMs by treating padding EOS as a part of response.}}

At the core of \textbf{$\rho$-\texttt{EOS}} lies a simple but critical hypothesis: \textit{masked diffusion language models implicitly estimate how much generation space they need for a given input, even before the sequence is fully denoised.} 
Importantly, this behavior is not an emergent artifact of our inference strategy, but is deeply rooted in the training engineering implementation of masked dLLMs.
In masked diffusion large language models such as LLaDA~\cite{nie2025llada}, padding \texttt{EOS} tokens are explicitly treated as part of the response sequence and are randomly masked during training. As a result, the model is repeatedly trained to jointly predict semantic tokens and \texttt{EOS} tokens under partial observation. This training paradigm forces the model to reason not only about \textbf{\emph{what}} content should be generated, but also \textbf{\emph{where}} meaningful content should end. Consequently, the prediction of \texttt{EOS} becomes intrinsically coupled with the model’s internal estimate of length sufficiency.

Crucially, this latent estimate is not exposed as an explicit scalar length prediction. Instead, it manifests implicitly through property of special functional token--the density of \texttt{EOS} over the remaining masked positions.
From a denoising perspective, when the model assigns high probability mass to \texttt{EOS} at masked locations, it signals that the semantic content is largely complete and that the remaining generation space is redundant.
Conversely, a consistently low implicit \texttt{EOS} density indicates that the model still requires additional capacity to express the intended response, and is therefore reluctant to terminate.
Viewed from this angle, \textbf{$\rho$-\texttt{EOS}} does not introduce an external heuristic for length control. Instead, it \textbf{\emph{reads out}} and \textbf{\emph{amplifies}} a signal that the model has already learned to encode during training.
By monitoring the implicit \texttt{EOS} density and feeding it back into the denoising process, our method aligns inference-time length adaptation with the model’s internal notion of response completeness.

This close alignment between training and inference is a key reason why \textbf{$\rho$-\texttt{EOS}} enables stable and effective bidirectional length control without additional supervision or multi-stage procedures.
\textbf{\emph{In essence, variable-length inference emerges naturally when the inference process respects the semantics of \texttt{EOS} prediction learned by masked diffusion large language models.}}

\subsection{\textbf{Why does implicit \texttt{EOS} density $\rho$ outperform \texttt{EOS} confidence for single-stage length control? -- A side effect of treating padding \texttt{EOS} as part of the response.}}

While treating padding \texttt{EOS} tokens as part of the response endows masked diffusion language models with implicit length awareness, it also introduces a non-trivial side effect that overconfidence on \texttt{EOS} may lead to instability in using it as a single-stage length control signal.

In particular, as observed in Yang et al.~\cite{yang2025taming}, masked dLLMs often exhibit an \texttt{EOS} Trap phenomenon: during early denoising steps, the predicted confidence of \texttt{EOS} is significantly higher than that of non-\texttt{EOS} ones. 
This behavior is a direct consequence of the training paradigm. Since padding \texttt{EOS} tokens may occupy a large fraction of the sequence during training and are randomly masked, the model is frequently optimized to recover \texttt{EOS} under high uncertainty.
As a result, in early decoding steps—when most positions remain masked and semantic structure has not yet emerged—the model tends to assign disproportionately high probability mass to \texttt{EOS}.
Importantly, this high \texttt{EOS} confidence does not indicate semantic completeness, but rather reflects the model’s uncertainty and its bias toward the statistically termination token.

When \texttt{EOS} confidence is directly used as a length control signal, this early-stage overconfidence can prematurely trigger contraction or suppress necessary expansion. Such misaligned length decisions are particularly harmful in a single-stage setting, where length adjustment and semantic denoising are tightly interleaved. This explains the performance degradation observed when \texttt{EOS} confidence is employed as the control signal for single-stage strategy.
In contrast, the implicit \texttt{EOS} density $\rho$ aggregates \texttt{EOS} mass over all remaining masked positions, providing a global and temporally smoother estimate of length sufficiency. Therefore, implicit density $\rho$ serves as a more robust length control signal than confidence.


\end{document}

%% file: sections/introduction.tex
\section{Introduction}
\label{sec:introduction}

Autoregressive large language models~\cite{brown2020language,bai2023qwen,touvron2023llama,achiam2023gpt4} have exhibited scaling laws and versatile capabilities in language understanding and generation. Recently, diffusion large language models (dLLMs)~\cite{nie2025llada,khanna2025mercury,geminidiffusion,song2025seed} have emerged as a promising alternative paradigm, attracting increasing attention.
In contrast to autoregressive models that generate next-tokens sequentially, dLLMs formulate text generation as a discrete probabilistic denoising process. Starting from a fully masked sequence, dLLMs iteratively unmask \texttt{MASK} tokens in parallel, which enables global context modeling and flexible generation orders.

\begin{figure*}[t!]
    \begin{center}
    \includegraphics[width=0.9\linewidth]{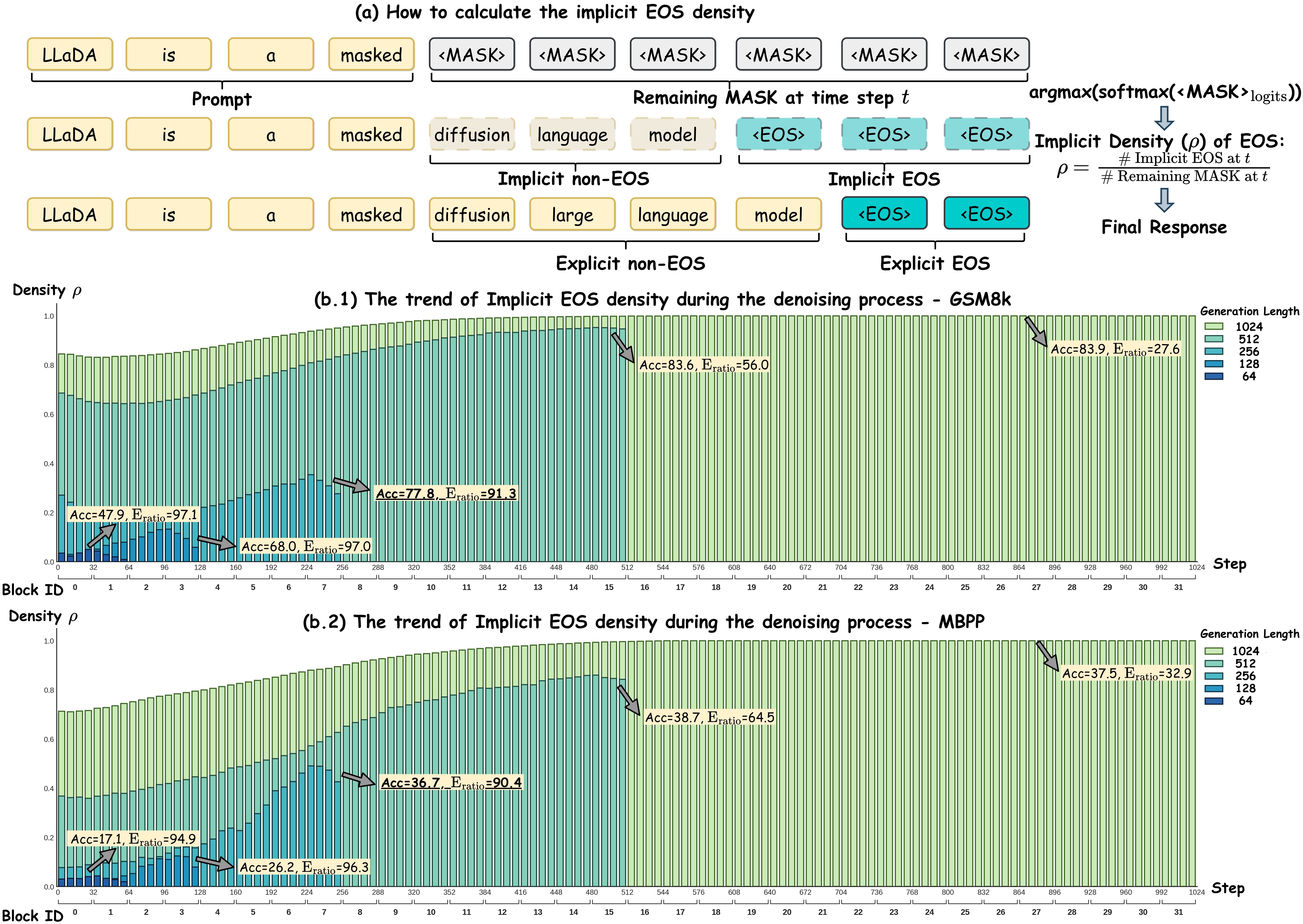}
    \end{center}
    \vspace{-1mm}
    \caption{\textbf{The Evolution Trend of Implicit \texttt{EOS} Density During the Denoising Process.} (a) Illustration of the calculation for implicit \texttt{EOS} density. (b.1) Trends of implicit \texttt{EOS} density on GSM8k. (b.2) Trends of implicit \texttt{EOS} density on MBPP.}
    \label{fig:motivation}
    \vspace{-4mm}
\end{figure*}

Despite these advantages, the generation length of dLLMs is fundamentally constrained by their length-rigid generation paradigm. Specifically, inference starts from a fixed number of pre-filled \texttt{MASK} tokens, which strictly determines the final output length.
In practice, empirically specifying this length presents a severe dilemma, particularly for long chain reasoning~\cite{lightman2023mat5h00,muennighoff2025s1} or tasks that require adaptive switching between short and long thinking modes (hybrid reasoning)~\cite{claude4systemcard2025,claude4.5systemcard2025}.
If the predefined length is too short, the model lacks sufficient token budget to solve complex reasoning tasks. Conversely, adopting a universally long generation length incurs substantial computational overhead, and often results in a low token utilization for tasks with inherently short length, causing significant computational waste.
This rigidity stands in sharp contrast to autoregressive models, which can naturally adjust their output length at test time.

To mitigate this limitation, DAEDAL~\cite{li2025beyond} proposes a two-stage solution that supports only unidirectional expansion through separate initial length adjustment and iterative mask insertion phases.
Its first stage iteratively expands an initial short sequence to a coarse target length by checking trailing \texttt{EOS} tokens confidence: if the confidence of \texttt{EOS} tokens is below a threshold, a fixed block of \texttt{MASK} tokens is appended. The second stage then performs iterative mask insertion during denoising to further expand remaining low-confidence regions. 
While effective, this two-stage pipeline inevitably introduces additional inference latency, making it less suitable for high-throughput or latency-sensitive deployment scenarios.
In addition, DAEDAL only supports unidirectional length expansion, lacking the ability to contract the generation length, which becomes problematic when the initial length is overly long, or when earlier length expansion decisions are overly aggressive and require later contraction.

To this end, we revisit the denoising dynamics of masked dLLMs and identify an implicit length-indicative signal.
As illustrated in Figure~\ref{fig:motivation}, under different fixed generation lengths, the implicit \texttt{EOS} density exhibits clearly distinguishable trends throughout the denoising process.
When the implicit \texttt{EOS} density is very low, it indicates that the remaining length is insufficient, as the model requires additional space to generate semantic content. In contrast, when the density becomes high, it indicates that the remaining length is excessive, i.e., there is surplus generation capacity. When the length is appropriate---balancing performance and effective token ratio---the density gradually converges to an intermediate equilibrium region.
Crucially, this signal emerges naturally during denoising and reflects the model’s internal assessment of length sufficiency.
Motivated by this observation, we introduce \textbf{$\rho$-\texttt{EOS}}, a training-free and bidirectional variable-length denoising strategy. Unlike prior two-stage approaches, \textbf{$\rho$-\texttt{EOS}} enables denoising and length adaptation jointly within a single stage by monitoring the implicit \texttt{EOS} density.

Extensive experiments demonstrate that \textbf{$\rho$-\texttt{EOS}} achieves performance comparable to fixed-length baselines and DAEDAL with significantly improving inference efficiency and token utilization across diverse tasks.
Furthermore, compared to DAEDAL, our \textbf{$\rho$-\texttt{EOS}} supports bidirectional length adjustment, making it more flexible and effective for practical deployment scenarios.

%% file: sections/related_work.tex
\section{Related Work}
\label{sec:related work}

\begin{figure*}[t!]
    \begin{center}
    \includegraphics[width=0.99\linewidth]{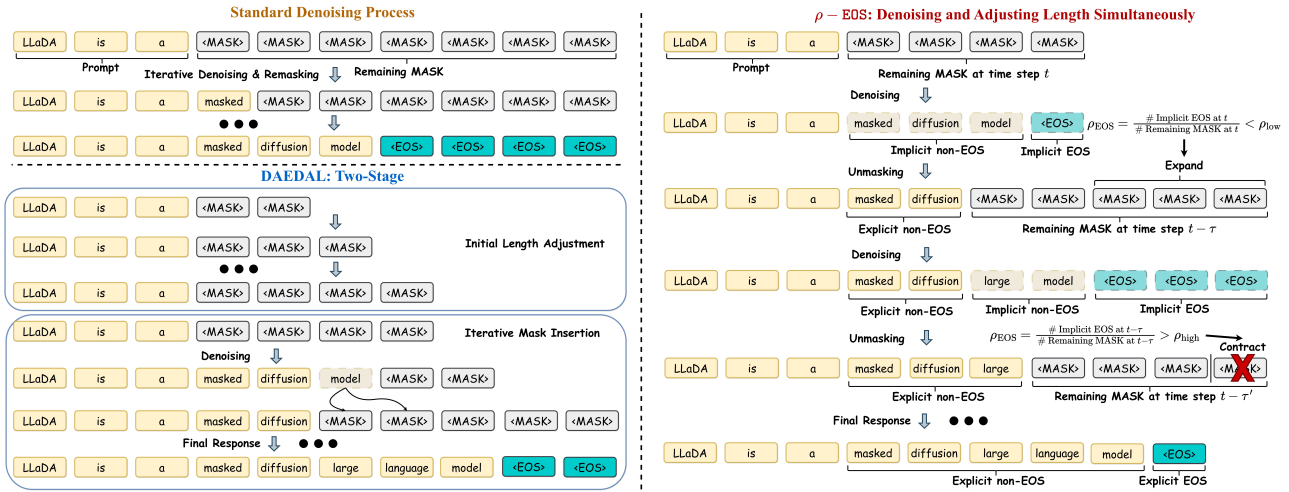}
    \end{center}
    \vspace{-1mm}
    \caption{\textbf{The Formulation of Implicit and Explicit Concept for Token and Density.} \textbf{(Upper Left)} The standard denoising process of masked diffusion large language models (e.g., LLaDA). \textbf{(Lower Left)} The two stage of DAEDAL. \textbf{(Right)} The bidirectional variable-length denoising process of \textbf{$\rho$-\texttt{EOS}}.}
    \label{fig:rho_EOS}
    \vspace{-4mm}
\end{figure*}

\textbf{Diffusion Large Language Models.}
Given the success of diffusion models in continuous generative modeling, recent work has extended this approach to discrete domains for text generation~\cite{austin2021structured,he2023diffusionbert,gong2022diffuseq,lou2023discrete,zheng2024masked,sahoo2024simple}. A prominent paradigm is masked diffusion large language models (dLLMs), which iteratively denoise a masked sequence for bidirectional, parallel generation~\cite{nie2025llada,zhu2025llada15,zhu2025llada,gong2024scaling,dream2025}. However, this generation paradigm requires a predefine a fixed generation length, which can lead to either redundant computation when the predefined length is overly long or degraded generation quality when the length budget is insufficient~\cite{li2025beyond}.
Another line of dLLMs, known as block diffusion language models~\cite{arriola2025block,cheng2025sdar,liu2025wedlm}, naturally supports variable-length generation for adopting a hybrid paradigm that divides a sequence into blocks, performs discrete diffusion within each block, and conditions autoregressively on all preceding blocks.
In this work, we focus on masked dLLMs and aim to address fixed-length limitation without compromising single-stage denoising paradigm.

\textbf{Variable-Length Strategies for Masked dLLMs.} 
To alleviate the fixed-length constraint of masked dLLMs, recent works have explored training-based and training-free approaches.
For instance, FlexMDMs~\cite{kim2025any} decompose the denoising process into insertion and unmasking steps, learning to predict the number of new \texttt{MASK} tokens to insert before each position. 
DreamOn~\cite{Dreamon2025} introduces special control tokens, where \texttt{EXPAND} splits into two \texttt{MASK} tokens to extend the sequence, and \texttt{DELETE} is removed to shorten it. 
dLLM-Var~\cite{yang2025diffusion} enables masked dLLMs be able to infer in a block diffusion manner via training the models to accurately predict the \texttt{EOS} token.
In contrast, DAEDAL~\cite{li2025beyond} proposes a training-free alternative that decouples length adjustment from denoising. It first performs length expansion then follow a adaptive denoising. While DAEDAL improves inference efficiency to some extent, its two-stage design potentially introduces additional latency and only supports unidirectional length expansion, which limits flexibility.
Moreover, length flexibility also plays a critical role in reinforcement learning (RL) fine-tuning, where diverse reasoning depths and response structures can facilitate exploration~\cite{zhao2025d1,gong2025diffucoder,yang2025taming,wang2025spg}. In such settings, RL rollouts and high-throughput deployment are often tightly bottlenecked by inference latency, making multi-stage decoding strategies particularly unfavorable.
These considerations further motivate the need for a single-stage, training-free, and bidirectional length control mechanism tailored to masked dLLMs.

%% file: sections/methods.tex
\section{Methods}
\label{sec:methods}

\subsection{Overview}
\label{subsec:overview}
To address the fixed-length limitation in standard masked diffusion large language models (dLLMs) inference, we introduce \textbf{$\rho$-\texttt{EOS}}, a training-free, single-stage bidirectional variable-length denoising strategy that enables dLLMs to dynamically adapt their generation length \emph{on the fly} during the denoising process itself, rather than predefining fixed-length or relying on multi-stage heuristics adjustments.

\subsection{Implicit Token}
\label{subsec:implicit tokens}
We first clarify the concept of \textbf{Implicit Token}. At each denoising step, the input $\mathbf{x}_t$ at time step $t$ is forwarded through the model ($p_{\theta}(\cdot)$) to obtain the logits of input (denote as $\mathbf{x}_t^{\text{logits}}$). These logits are then processed via $\mathrm{argmax}(\mathrm{softmax}(\cdot))$, yielding a prediction of tokens:
\begin{align}
\mathbf{x}_t^\text{logits} & = p_{\theta}(\mathbf{x}_t), \\
\hat{\mathbf{x}}_0  & = \mathrm{argmax}(\mathrm{softmax}(\mathbf{x}_t^\text{logits})), \\
\mathbf{x}^{t-\tau}  & = \text{Remask}(\hat{\mathbf{x}}_0),
\label{eq:implicit token}
\end{align}
where $p_{\theta}(\cdot)$ denotes the model, and $\hat{\mathbf{x}}_0$ represents the model's prediction of the clean sequence. We refer to $\hat{\mathbf{x}}_0$ as the implicit $\mathbf{x}_0$, since it is not immediately committed as final response. 
Moreover, the tokens in $\hat{\mathbf{x}}_0$ that correspond to the remaining \texttt{MASK} position in $\mathbf{x}_t$ are defined as the \textbf{Implicit Tokens} at time step $t$ (i.e., $\{i|\hat{\mathbf{x}}^i_0 \land (\mathbf{x}^i_t=\texttt{MASK})\}$).
After applying $\text{Remask}(\cdot)$, only a subset of $\hat{\mathbf{x}}_0$ is decoded into \textbf{Explicit Tokens}, while the remaining implicit tokens of $\hat{\mathbf{x}}_0$ are remasked to form $\mathbf{x}_{t-\tau}$.
This iterative process continues until a fully denoised sequence is obtained, which constitutes the standard inference procedure of masked dLLMs.
A schematic illustration is shown in Figure~\ref{fig:rho_EOS}.

\subsection{Implicit EOS Density}
\label{subsec:implicit EOS density}
Based on the notion of implicit tokens, we define \textbf{Implicit Density} ($\rho$) \textit{as the ratio between implicit tokens with specific properties and masked tokens at time step t}.
As depicted at the right panel of Figure~\ref{fig:rho_EOS}, implicit token density:
\begin{equation}
\label{eq:implicit density}
\rho  = \frac{\text{\# Implicit Token at}~t}{\text{\# Remaining \texttt{MASK} Token at}~t},
\end{equation}
where $\text{\# Implicit Token at}~t$ denotes the number of implicit tokens with certain properties at timestep $t$, and $\text{\# Remaining MASK Token at}~t$ represents the number of remaining \texttt{MASK} tokens in $\mathbf{x}_t$ needed to be denoised.
Subsequently, we conclude the implicit $\texttt{EOS}$ density:
\begin{equation}
\label{eq:implicit EOS density}
\rho_{\texttt{EOS}}  = \frac{\text{\# Implicit \texttt{EOS} Token at}~t}{\text{\# Remaining \texttt{MASK} Token at}~t}.
\end{equation}

\begin{figure}[t!]
    \begin{center}
    \includegraphics[width=1.\linewidth]{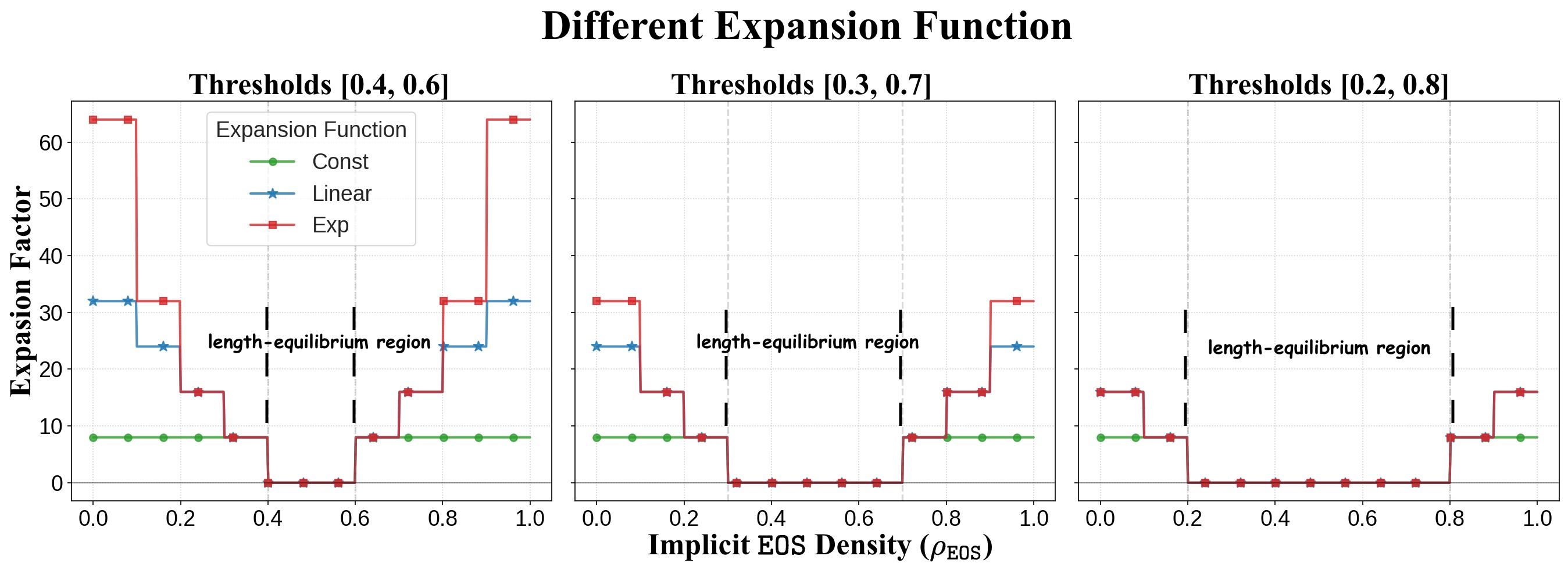}
    \end{center}
    \vspace{-3mm}
    \caption{Example of piecewise expansion factor function for constants, linearity, and exponentiation.}
    \label{fig:expansion factor function}
    \vspace{-3mm}
\end{figure}

\input{algorithms/rho_EOS}

\subsection{Bidirectional Length Adjustment}
\label{subsec:bidirectional length adjustment}
As illustrated in Figure~\ref{fig:motivation} (which shows the trend of $\rho_\texttt{EOS}$ on GSM8k and MBPP across fixed-length ranging from 64 to 1024), we observe that masked dLLMs implicitly encode length sufficiency information in $\rho_{\texttt{EOS}}$. When the masked budget is sufficient to complete the response, the model assigns higher probability mass to \texttt{EOS}, resulting in a larger $\rho_\texttt{EOS}$. Conversely, when the available length is insufficient, the model tends to fully utilize the masked space for semantic contents, leading to a persistently low $\rho_{\texttt{EOS}}$.
This behavior indicates that $\rho_\texttt{EOS}$ serves as a reliable internal signal for assessing whether the current generation length is appropriate.

Similarly, as demonstrated in Figure~\ref{fig:motivation}, when the generation length is excessively long (e.g., 512 or 1024), the $ \rho_\texttt{EOS}$ converges toward near 1, indicating a waste of capacity. Conversely, when the length is too short (e.g., 64 or 128), the density converges toward 0, signaling insufficient semantic content.
This observation suggests that the ideal implicit \texttt{EOS} density does not converge to an extreme value but rather reaches an intermediate \textbf{length-sufficiency equilibrium state}, which reflects a suitable length for both quality and efficiency. 
To operate this criterion robust, we empirically define the \textbf{length-equilibrium region} using two thresholds $\rho_\text{low}$ and $\rho_\text{high}$. When $\rho_\texttt{EOS} \in [\rho_\text{low},\rho_\text{high}]$, the model suspends length modification and focuses solely on denoising, enabling stable bidirectional control without oscillatory behavior.
When $\rho_{\texttt{EOS}} < \rho_\text{low}$, the remaining masked space is insufficient and length expansion is triggered via \texttt{MASK} insertion.
Conversely, when $\rho_\texttt{EOS} > \rho_\text{high}$, the model indicates surplus generation capacity and the sequence is contracted by removing partial trailing \texttt{MASK}.

\input{tables/main_results_0}

Formally, the length adjustment is defined as:
\begin{align}
\label{eq:x adjustment}
\mathbf{x} = 
& \left\{
\begin{aligned} 
& \mathbf{x} \leftarrow \mathbf{x} \oplus [\underbrace{\texttt{MASK}, \dots, \texttt{MASK}}_{\mathrm{E}_{\text{factor}}(\rho_{\texttt{EOS}})}] \quad ~~ \text{if  } \rho_\texttt{EOS} < \rho_\text{low}\\ 
& \mathbf{x} \leftarrow \mathbf{x}^{0:(|\mathbf{x}|-\mathrm{E}_{\text{factor}}(\rho_{\texttt{EOS}}))}  \quad \qquad \text{else if  } \rho_\texttt{EOS} > \rho_\text{high} \\ 
& \mathbf{x}  \quad \qquad \qquad \qquad \qquad \qquad \text{otherwise  } \\ 
\end{aligned}
\right.
\end{align}
with the update to current generation length $L_\text{cur}$:
\begin{align}
\label{eq:length adjustment}
L_\text{cur} = 
& \left\{
\begin{aligned} 
& L_\text{cur} + \mathrm{E}_\text{factor}(\rho_\texttt{EOS}) \qquad ~~ \text{if  } \rho_\texttt{EOS} < \rho_\text{low}\\ 
& L_\text{cur} - \mathrm{E}_\text{factor}(\rho_\texttt{EOS})  \qquad \text{else if  } \rho_\texttt{EOS} > \rho_\text{high} \\ 
& L_\text{cur}  \qquad \qquad \qquad \qquad ~ \text{otherwise  } \\ 
\end{aligned}
\right.
\end{align}
Beyond above, we further investigate how aggressively the adjustment should be performed through different \textbf{expansion factor functions} $\mathrm{E}_\text{factor}(\cdot)$. DAEDAL~\cite{li2025beyond} adopts a constant expansion factor, a design that is simple and stable but lead to multiple adjustment iterations when the initial length is severely mismatched.
Motivated by Figure~\ref{fig:motivation}, we additionally consider linear and exponential designs, as depicted in Figure~\ref{fig:expansion factor function}. 
The piecewise linear function scales the adjustment magnitude proportionally to the distance between $\rho_{\texttt{EOS}}$ and the length-equilibrium region boundaries.
The piecewise exponential function further amplifies this effect when $\rho_{\texttt{EOS}}$ is extremely low or high, enabling more aggressive expansion or contraction.
The key motivation is to rapidly drive the generation into the stable $\rho_{\texttt{EOS}}$ region, so that the model can enter the pure denoising phase early without repeatedly interleaving length adjustment steps.
Compared to unidirectional expansion, bidirectional control improves flexibility and token utilization. As illustrated in Figure~\ref{fig:rho_EOS}, if the length becomes overly aggressive after expansion, the proposed mechanism can actively contract the sequence, leading to more efficient and stable generation.

\subsection{One-Stage Bidirectional Variable-Length Denoising}
\label{subsec:one-stage bidirectional variable-length denoising}
Overall, by integrating implicit \texttt{EOS} density estimation with bidirectional length adjustment, \textbf{$\rho$-\texttt{EOS}} realizes \textbf{training-free} and \textbf{bidirectional variable-length} generation in a fully unified, single-stage denoising process, i.e., jointly performs unmasking (decoding) and length adjustment. 
At each denoising step, the model $p_{\theta}(\cdot)$ performs once forward pass on the $\mathbf{x}_t$ to obtain $\mathbf{x}_t^\text{logits}$ and the implicit prediction $\hat{\mathbf{x}}_0$ (Equation~\ref{eq:implicit token}).
After applying the $\mathrm{Remask}(\cdot)$ operation to produce $\mathbf{x}_{t-\tau}  = \mathrm{Remask}(\hat{\mathbf{x}}_0)$, we compute $\rho_{\texttt{EOS}}$ according to Equation~\ref{eq:implicit EOS density} and adjust the current sequence length using the bidirectional strategy in Equations~\ref{eq:x adjustment} and~\ref{eq:length adjustment}.
This step-level denoising process is repeated until entire denoising completes. By coupling denoising and length adjustment within a single inference loop, \textbf{$\rho$-\texttt{EOS}} enables dynamic expansion when additional reasoning space is needed and contraction when redundancy is detected, without introducing extra stages.
Algorithm~\ref{alg:rho_EOS} provides a detailed description of the full procedure.

%% file: algorithms/rho_EOS.tex
\begin{algorithm}[t]
\caption{The Inference Process of \textbf{$\rho$-\texttt{EOS}}}
\label{alg:rho_EOS}
\begin{algorithmic}[1]
\State \textbf{Input:} Prompt $\mathbf{p}$, model $p_{\theta}$, initial/max length $L_{\text{init}}/L_{\text{max}}$, density thresholds $[\rho_{\text{low}}, \rho_{\text{high}}]$, confidence threshold $\tau_\text{high}$, expansion factor function $\mathrm{E}_{\text{factor}}(\cdot)$, max length adjustment step $N$
\State \textbf{Output:} clean sequence $\mathbf{x}$

\State $L_{\text{cur}} \leftarrow L_{\text{init}}$, $n \leftarrow 0$
\State $\mathbf{x}^t \leftarrow [\mathbf{p}, \underbrace{\texttt{MASK}, \dots, \texttt{MASK}}_{L_{\text{init}}}]$ \Comment{Initialize sequence}
\While{$\text{ContainsMask}(\mathbf{x}_t)$}
    \State $\mathbf{x}_t^{\text{logits}} \leftarrow p_{\theta}(\mathbf{x}_t)$
    \State $\rho_{\texttt{EOS}} \leftarrow \text{ImplicitEOSDensity}(\mathbf{x}_t^{\text{logits}})$
    \State $\mathbf{x}_t^{\text{conf}}, \hat{\mathbf{x}}_0 \leftarrow \text{Decode}(\mathbf{x}_t^{\text{logits}})$

    \For{$i \in \{(\mathbf{x}_t^i = \texttt{MASK}) \land (\mathbf{x}_t^{\text{conf}, i} > \tau_\text{high})\}$}
        \State $\mathbf{x}_{t - \tau}^i \leftarrow \hat{\mathbf{x}}_0^i$
    \EndFor
    
    \If{$0<L_{\text{cur}}<L_{\text{max}}$ and $n<N$}
        \If{$\rho_{\texttt{EOS}} \in [\rho_{\text{low}}, \rho_{\text{high}}]$}
            \State $\mathbf{x}_t, L_{\text{cur}} \leftarrow \mathbf{x}_{t - \tau}, L_{\text{cur}}$ \Comment{Hold $L_{\text{cur}}$}
        \ElsIf{$\rho_{\texttt{EOS}} < \rho_{\text{low}}$} \Comment{Expand $L_{\text{cur}}$} 
                \State $\mathbf{x}_{t - \tau} \leftarrow \mathbf{x}_{t - \tau} \oplus [\underbrace{\texttt{MASK}, \dots, \texttt{MASK}}_{\mathrm{E}_{\text{factor}}(\rho_{\texttt{EOS}})}]$
                \State $L_{\text{cur}} \leftarrow L_{\text{cur}} + \mathrm{E}_{\text{factor}}(\rho_{\texttt{EOS}})$
        \Else \Comment{Contract $L_{\text{cur}}$}
                \State $\mathbf{x}_{t - \tau} \leftarrow \mathbf{x}_{t - \tau}^{0:(|\mathbf{x}|-\mathrm{E}_{\text{factor}}(\rho_{\texttt{EOS}}))}$ 
                \State $L_{\text{cur}} \leftarrow L_{\text{cur}} - \mathrm{E}_{\text{factor}}(\rho_{\texttt{EOS}})$
        \EndIf
    \EndIf
    
    \State $n \leftarrow n+1$
\EndWhile
\State \textbf{return} $\mathbf{x}$
\end{algorithmic}
\end{algorithm}

%% file: tables/main_results_0.tex
\begin{figure*}[t!]
    \centering
    \begin{minipage}[t]{0.69\textwidth}
        \centering
        \renewcommand{\arraystretch}{1.1}
        \begin{table}[H]
            \centering
            \caption{\textbf{Main Results on LLaDA-Instruct-8B across Four Benchmarks.} We compare the \textbf{$\rho$-\texttt{EOS}} performance against DAEDAL and various baseline configurations. $\boldsymbol{Acc}$ denotes accuracy, $\boldsymbol{E_\text{token}}$ is the average effective tokens (the response length excluding trailing padding), $\boldsymbol{N_\text{token}}$ is the average total tokens, and $\boldsymbol{E_\text{ratio}}$ is the effective token ratio. $\boldsymbol{T_\text{runtime}}$ represents the runtime spent on the evaluation (in seconds). The best configuration for the baseline is highlighted in \textcolor{baseline_orange}{\textbf{orange}}. DAEDAL is highlighted in \textcolor{daedal_blue}{\textbf{blue}}, and \textbf{$\rho$-\texttt{EOS}} is highlighted in \textcolor{rho_EOS_red}{\textbf{red}}. Under the \textbf{$\rho$-\texttt{EOS}} setting, Sym and Asym denote the symmetric and asymmetric lower-upper threshold centered around 0.5. The \textbf{best} results are \textbf{bold}, and the \underline{second-best} results are \underline{underlined}.}
            \label{tab:main_results_0}
            \vspace{-0.1cm}
            \resizebox{0.99\textwidth}{!}{
            \begin{tabular}{llccccccccc}
                \toprule
                \multirow{2}{*}{\textbf{Benchmark}} & \multirow{2}{*}{\textbf{Metric}} & \multicolumn{6}{c}{\textbf{Fixed-Length Denoising (Baseline)}} & \multicolumn{1}{c}{\cellcolor{lightblue}\textbf{DAEDAL}} & {\cellcolor{lightred}\textbf{$\rho$-\texttt{EOS}}} & {\cellcolor{lightred}\textbf{$\rho$-\texttt{EOS}}} \\
                \cmidrule(lr){3-11}
                & & 64 & 128 & 256 & 512 & 1024 & 2048 & \multicolumn{1}{c}{\cellcolor{lightblue}64} & \multicolumn{1}{c}{\cellcolor{lightred}64 Sym} & \multicolumn{1}{c}{\cellcolor{lightred}64 Asym} \\
                \midrule
                \multirow{5}{*}{\textbf{GSM8K}} & $\boldsymbol{Acc}$ & 47.9 & 68.0 & 77.8 & 83.6 & \cellcolor{lightorange}{83.9} & 82.7 & \cellcolor{lightblue}\textbf{84.6} & \cellcolor{lightred}{81.8} & \cellcolor{lightred}\underline{84.2} \\
                & $\boldsymbol{E_\text{token}}$ & 62.1 & 124.1 & 233.7 & 286.8 & \cellcolor{lightorange}{282.9} & 297.3 & \cellcolor{lightblue}{246.0} & \cellcolor{lightred}{198.4} & \cellcolor{lightred}{252.5} \\
                & $\boldsymbol{N_\text{token}}$ & 64 & 128 & 256 & 512 & \cellcolor{lightorange}{1024} & 2048 & \cellcolor{lightblue}{331.0} & \cellcolor{lightred}{231.8} & \cellcolor{lightred}{361.0} \\
                & $\boldsymbol{E_\text{ratio}}$ & 97.1\% & 97.0\% & 91.3\% & 56.0\% & \cellcolor{lightorange}{27.6\%} & 14.5\% & \cellcolor{lightblue}\underline{74.5\%} & \cellcolor{lightred}\textbf{85.6\%} & \cellcolor{lightred}{70.0\%} \\
                & $\boldsymbol{T_\text{runtime}}$ & 174 & 360 & 860 & 2458 & \cellcolor{lightorange}{8238} & 30305 & \cellcolor{lightblue}{1090} & \cellcolor{lightred}\textbf{645} & \cellcolor{lightred}\underline{823} \\
                \midrule
                \multirow{5}{*}{\textbf{MATH500}} & $\boldsymbol{Acc}$ & 24.0 & 29.0 & 35.4 & 38.4 & \cellcolor{lightorange}\underline{40.4} & 40.0 & \cellcolor{lightblue}{40.0} & \cellcolor{lightred}{37.0} & \cellcolor{lightred}\textbf{40.6} \\
                & $\boldsymbol{E_\text{token}}$ & 61.7 & 123.3 & 244.7 & 423.9 & \cellcolor{lightorange}{585.8} & 706.5 & \cellcolor{lightblue}{464.6} & \cellcolor{lightred}{428.8} & \cellcolor{lightred}{558.7} \\
                & $\boldsymbol{N_\text{token}}$ & 64 & 128 & 256 & 512 & \cellcolor{lightorange}{1024} & 2048 & \cellcolor{lightblue}{618.3} & \cellcolor{lightred}{486.3} & \cellcolor{lightred}{682.2} \\
                & $\boldsymbol{E_\text{ratio}}$ & 96.4\% & 96.3\% & 95.6\% & 82.8\% & \cellcolor{lightorange}{57.2\%} & 34.5\% & \cellcolor{lightblue}{75.2\%} & \cellcolor{lightred}\textbf{88.2\%} & \cellcolor{lightred}\underline{81.9\%} \\
                & $\boldsymbol{T_\text{runtime}}$ & 98 & 208 & 462 & 1202 & \cellcolor{lightorange}{3663} & 12764 & \cellcolor{lightblue}{2207} & \cellcolor{lightred}\textbf{1398} & \cellcolor{lightred}\underline{1968} \\
                \midrule
                \multirow{5}{*}{\textbf{MBPP}} & $\boldsymbol{Acc}$ & 21.0 & 28.8 & 36.7 & 38.7 & 37.5 & \cellcolor{lightorange}{38.8} & \cellcolor{lightblue}{39.4} & \cellcolor{lightred}\underline{39.6} & \cellcolor{lightred}\textbf{40.6} \\
                & $\boldsymbol{E_\text{token}}$ & 61 & 122 & 232 & 331 & 335 & \cellcolor{lightorange}{336} & \cellcolor{lightblue}{308.4} & \cellcolor{lightred}{324.0} & \cellcolor{lightred}{335.9} \\
                & $\boldsymbol{N_\text{token}}$ & 64 & 128 & 256 & 512 & 1024 & \cellcolor{lightorange}{2048} & \cellcolor{lightblue}{558.3} & \cellcolor{lightred}{507.4} & \cellcolor{lightred}{594.4} \\
                & $\boldsymbol{E_\text{ratio}}$ & 94.9\% & 95.5\% & 90.4\% & 64.5\% & 32.9\% & \cellcolor{lightorange}{16.3\%} & \cellcolor{lightblue}{55.2\%} & \cellcolor{lightred}\textbf{63.9\%} & \cellcolor{lightred}\underline{55.5\%} \\
                & $\boldsymbol{T_\text{runtime}}$ & 176 & 332 & 713 & 1728 & 4760 & \cellcolor{lightorange}{15172} & \cellcolor{lightblue}{3551} & \cellcolor{lightred}\underline{2031} & \cellcolor{lightred}\textbf{2030} \\
                \midrule
                \multirow{5}{*}{\textbf{HUMANEVAL}} & $\boldsymbol{Acc}$ & 17.1& 26.2 & 36.6& 45.1& 45.1 & \cellcolor{lightorange}\textbf{48.2} & \cellcolor{lightblue}\underline{44.5} & \cellcolor{lightred}{43.3} & \cellcolor{lightred}{43.9} \\
                & $\boldsymbol{E_\text{token}}$ & 59.6 & 125 & 247.3 & 462 & 631.1 & \cellcolor{lightorange}{691.1} & \cellcolor{lightblue}{494.7} & \cellcolor{lightred}{570.4} & \cellcolor{lightred}{643.1} \\
                & $\boldsymbol{N_\text{token}}$ & 64 & 128 & 256 & 512 & 1024 & \cellcolor{lightorange}{2048} & \cellcolor{lightblue}{813.0} & \cellcolor{lightred}{654.0} & \cellcolor{lightred}{769.9} \\
                & $\boldsymbol{E_\text{ratio}}$ & 93.2\% & 97.7\% & 96.6\% & 90.2\% & 61.6\% & \cellcolor{lightorange}{33.8\%} & \cellcolor{lightblue}{64.8\%} & \cellcolor{lightred}\textbf{88.4\%} & \cellcolor{lightred}\underline{83.5\%} \\
                & $\boldsymbol{T_\text{runtime}}$ & 111 & 230 & 543 & 1474 & 4569 & \cellcolor{lightorange}{16046} & \cellcolor{lightblue}{1283} & \cellcolor{lightred}\textbf{580} & \cellcolor{lightred}\underline{593} \\
                \bottomrule
            \end{tabular}
            }
            \end{table}
    \end{minipage}
    \begin{minipage}[t]{0.305\textwidth}
    \vspace{12pt}
        \centering
        \includegraphics[width=0.88\textwidth]{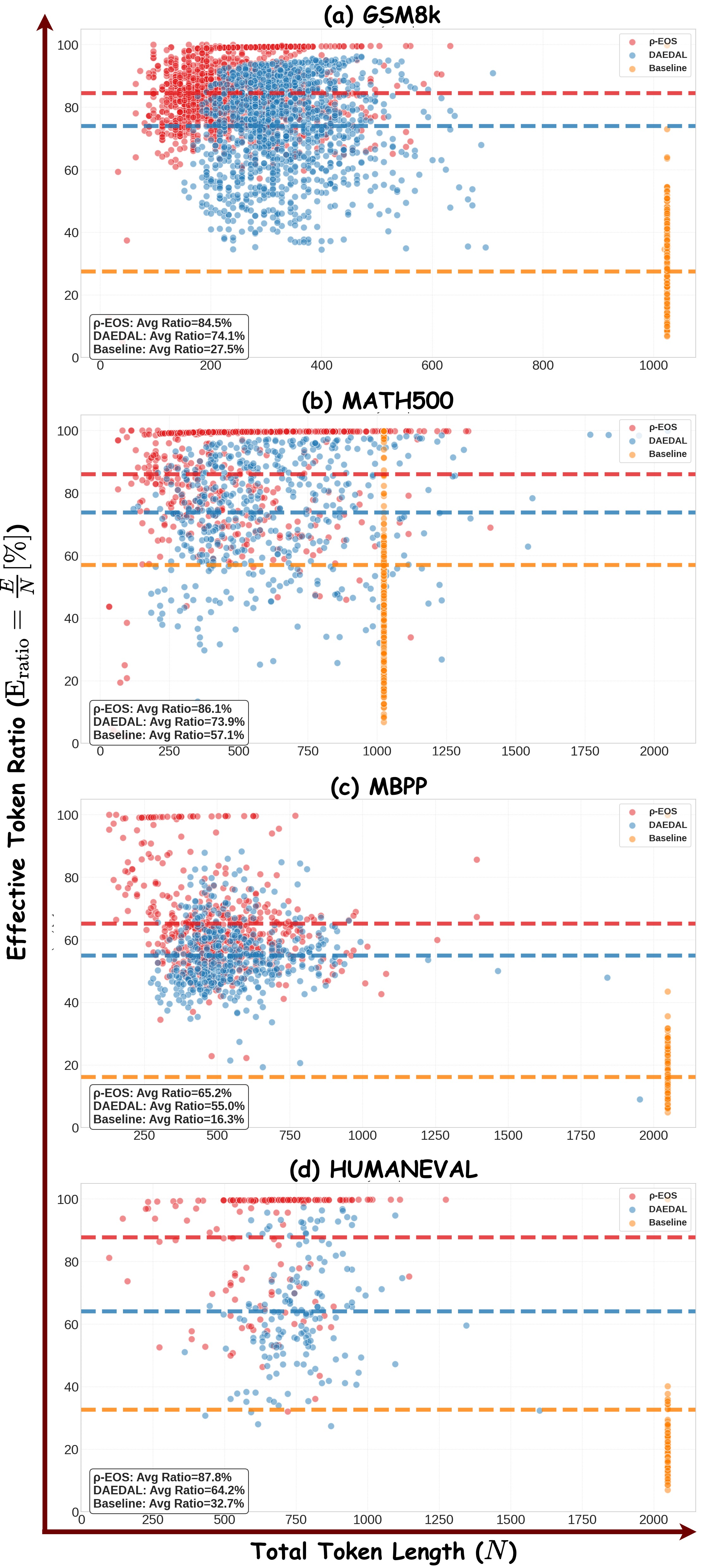}
        \vspace{-0.2cm}
        \caption{$\mathrm{E}_\text{ratio}$ per sample.} 
        \label{fig:LLaDA_E_ratio}
    \end{minipage}
\vspace{-0.4cm}
\end{figure*}

%% file: sections/experiments.tex
\section{Experiments}
\label{sec:experiments}

\input{tables/main_results_1}

\subsection{Experimental Setups}
\label{subsec:experimental setups}
\textbf{Implementation Details.}
We follow the experimental settings of DAEDAL~\cite{li2025beyond} to ensure fairness and reproducibility. All experiments are conducted under the standard generation, without employing additional acceleration or caching mechanisms, and the batch size is set to 8. To mitigate randomness from a single evaluation run, each experiment is repeated three times, and we report the averaged results.

\textbf{Benchmarks and Metrics.}
To comprehensively evaluate the effectiveness of \textbf{$\rho$-\texttt{EOS}}, we conduct experiments on four benchmarks covering mathematics and code task.
For mathematics, we evaluate on GSM8K~\citep{cobbe2021gsm8k}, which consists of grade-school math word problems, and MATH500\citep{lightman2023mat5h00}, a more challenging benchmark composed of competition-level mathematical problems. Accuracy is adopted as the performance metric.
For code generation, we employ MBPP\citep{austin2021mbpp}, which focuses on entry-level Python programming tasks, and HumanEval\citep{chen2021humaneval}, a more challenging handwritten benchmark for program synthesis. We report the standard pass@1 (accuracy) metric to assess functional correctness. In addition to accuracy ($\boldsymbol{Acc}$), we introduce four efficiency-related metrics: the total number of generated tokens ($\boldsymbol{N_\text{token}}$), the number of effective tokens after removing trailing \texttt{EOS} padding ($\boldsymbol{E_\text{token}}$), the effective token ratio ($\boldsymbol{E_\text{ratio}=\frac{E_\text{token}}{N_\text{token}}}$), and the evaluation runtime ($\boldsymbol{T_\text{runtime}}$, in seconds).

\input{tables/var_init_len}
\input{tables/ablation_rho}

\subsection{Main Results}
\label{subsec:main results}
We report the main results in Table~\ref{tab:main_results_0} and Table~\ref{tab:main_results_1}, comparing fixed-length baselines, DAEDAL, and our proposed \textbf{$\rho$-\texttt{EOS}}. The fixed-length baselines are evaluated under six predefined generation lengths ranging from 64 to 2048. Both DAEDAL and \textbf{$\rho$-\texttt{EOS}} start from the same short initial length, $L_\text{init}=64$, ensuring a fair comparison.

\textbf{Performance of $\rho$-\texttt{EOS} with an initial length of 64.} 
As shown in Table~\ref{tab:main_results_0} and Table~\ref{tab:main_results_1}, \textbf{$\rho$-\texttt{EOS}} achieves performance comparable to DAEDAL and the best-performing fixed-length baselines across most benchmarks. Under MATH500 and MBPP in Table~\ref{tab:main_results_0}, the performance of \textbf{$\rho$-\texttt{EOS}} with $L_\text{init}=64$ Asym suppresses both of them, and \textbf{$\rho$-\texttt{EOS}} consistently demonstrates substantially higher efficiency.
Under MBPP and HUMANEVAL in Table~\ref{tab:main_results_1}, the performance of \textbf{$\rho$-\texttt{EOS}} with $L_\text{init}=64$ Asym suppresses DAEDA, and \textbf{$\rho$-\texttt{EOS}} consistently outperforms DAEDAL and best-performance baseline on efficiency.
More importantly, both the effective token ratio ($\boldsymbol{E_\text{ratio}}$) and evaluation runtime ($\boldsymbol{T_\text{runtime}}$) significantly outperform DAEDAL and the strongest fixed-length baselines on nearly all tasks.
By simultaneously performing length adjustment and denoising within a unified diffusion process, \textbf{$\rho$-\texttt{EOS}} adapts task-dependent generation lengths on the fly, achieving competitive accuracy at a substantially lower computational cost.

\textbf{$\rho$-\texttt{EOS} significantly improves effective token ratio and evaluation speed.} 
For best-performance fixed-length baselines, the price for this performance comes at the cost of sharply increased evaluation runtime ($\boldsymbol{T_\text{runtime}}$) and significantly reduced effective token ratio ($\boldsymbol{E_\text{ratio}}$). 
In particular, configurations such as (MBPP, 2048), (HUMANEVAL, 2048) in Table~\ref{tab:main_results_0}, and (GSM8k, 2048) in Table~\ref{tab:main_results_1} exhibit extremely low $\boldsymbol{E_\text{ratio}}$, accompanied by a dramatic increase in $\boldsymbol{T_\text{runtime}}$.
While DAEDAL partially alleviates this inefficiency, \textbf{$\rho$-\texttt{EOS}} further improves computational efficiency. 
By enabling bidirectional length adjustment, \textbf{$\rho$-\texttt{EOS}} consistently achieves higher $\boldsymbol{E_\text{ratio}}$ than both DAEDAL and the best-performing fixed-length baselines at comparable accuracy levels.
This reduces unnecessary bidirectional attention over excessively long sequences and minimizes wasted computation on redundant padding tokens.
To provide a more fine-grained view, Figure~\ref{fig:LLaDA_E_ratio} and Figure~\ref{fig:LLaDA_1p5_E_ratio} visualize the per-sample distribution of total generation lengths ($\boldsymbol{N_\text{token}}$) and the corresponding effective token ratio $\boldsymbol{E_\text{ratio}}$.
Across all four benchmarks, a substantial portion of samples generated by \textbf{$\rho$-\texttt{EOS}} achieve an $\boldsymbol{E_\text{ratio}}$ close to 100\%, approaching the efficiency of autoregressive counterparts.
Such behavior is rarely observed in DAEDAL or fixed-length baselines, further illustrating the superior efficiency of variable-length denoising with bidirectional control.

\subsection{Robustness and Ablation Studies}
\label{subsec:robustness and ablation studies}
\textbf{Robustness to the initial length}. 
Table~\ref{tab:var_init_len} demonstrates that \textbf{$\rho$-\texttt{EOS}} exhibits strong robustness to the choice of the initial generation length. When varying $L_\text{init}$ from 128 to 1024, \textbf{$\rho$-\texttt{EOS}} maintains remarkably stable accuracy, while consistently achieving higher effective token ratio and significantly lower runtime compared to DAEDAL across all benchmarks. In contrast, DAEDAL shows a clear sensitivity to the initial length on $\boldsymbol{E}_\text{ratio}$ and $\boldsymbol{T}_\text{runtime}$ aspects.
The robustness of stems from its \textbf{bidirectional length adjustment mechanism driven by implicit \texttt{EOS} density}. Rather than committing to a fixed or unidirectional expansion strategy, \textbf{$\rho$-\texttt{EOS}} dynamically contracts or expands the sequence based on whether the current length is over- or under-sufficient. As a result, even when initialized with an aggressive or conservative length, the model can rapidly converge toward a task-adaptive length regime. This explains why \textbf{$\rho$-\texttt{EOS}} consistently achieves competitive or superior accuracy with reduced runtime across all initial length settings.

\input{tables/ablation_E_factor}

\input{tables/rho_vs_confidence}

\textbf{Ablation on $\rho$ thresholds [$\rho_\text{low}$, $\rho_\text{high}$]}. 
We further investigate the sensitivity of \textbf{$\rho$-\texttt{EOS}} to the choice of the density thresholds $[\rho_\text{low}, \rho_\text{high}]$. As shown in Table~\ref{tab:ablation_rho}, Sym (i.e., symmetrically setting $\rho_\text{low}$ and $\rho_\text{high}$ thresholds at intervals of 0.1) varying the thresholds over a wide range leads to only minor fluctuations in accuracy and runtime, indicating low sensitivity to precise threshold tuning.
Across different threshold configurations, \textbf{$\rho$-\texttt{EOS}} consistently maintains stable accuracy while exhibiting predictable trade-offs between effective token ratio and total token count. In particular, wider length-equilibrium regions (e.g., $[0.2,0.8]$) favor conservative length adjustment, resulting in higher $\boldsymbol{E}_\text{ratio}$ but slightly reduced $\boldsymbol{E}_\text{token}$ and $\boldsymbol{T}_\text{runtime}$. Conversely, narrower or asymmetric settings allow more aggressive adjustment, increasing $\boldsymbol{E}_\text{token}$ at the cost of additional denoising steps.
This robustness suggests that the \textbf{implicit \texttt{EOS} density provides a reliable and well-calibrated signal of length sufficiency}, making \textbf{$\rho$-\texttt{EOS}} resilient to moderate variations in hyper-parameter choices.
Moreover, the asymmetric threshold pair [0.4, 0.8] reach a peak performance when $L_\text{init}$=64. We attribute this to the fact that the current model starts from a shorter initial length of 64, so a prior, the model tends to become longer, requiring a loose lower bound and a tight upper bound. Of course, we also found in the experiment of longer initial lengths, they do not have this prior knowledge, so they may not need this asymmetric boundary.

\textbf{Ablation to the expansion factor function $\mathrm{E}_{\text{factor}}(\cdot)$.} 
Table~\ref{tab:ablation_rho} compares different selection of the expansion factor function $\mathrm{E}_{\text{factor}}(\cdot)$.
The constant expansion strategy in DAEDAL, improves over the baseline but remains limited in its responsiveness to large length mismatches. In contrast, the linear expansion function adjusts the expansion magnitude proportionally to the deviation of $\rho_\texttt{EOS}$ from the length-equilibrium region, enabling faster correction when the current length is moderately misaligned.
The exponential expansion function further amplifies this effect. When $\rho_\texttt{EOS}$ is extremely low or high, the strategy applies substantially stronger length adjustments, allowing it to rapidly reach a stable denoising regime. As evidenced by the higher accuracy and few average adjustment steps in Table~\ref{tab:ablation_rho}, these results confirm that \textbf{adapting the adjustment magnitude to the confidence of the length signal is crucial}, and that aggressive correction in early stages can significantly accelerate convergence without sacrificing accuracy.

\textbf{Comparison on Using Confidence and Density of \texttt{EOS} as Length Control Signal and Single Stage v.s. Two Stage.}
Table~\ref{tab:rho_vs_confidence} compares using \texttt{EOS} confidence versus \texttt{EOS} density ($\rho$) as the length control signal, under both single-stage and two-stage length adjustment strategies. Across both GSM8K and MBPP, \textbf{$\rho$-\texttt{EOS} almost consistently outperforms confidence-based control}.
In the single-stage setting, confidence-based signals suffer from instability, leading to suboptimal accuracy and inefficient length adjustment. 
As mentioned in Yang et al.~\cite{yang2025taming}, masked dLLMs face \texttt{EOS} Trap during early decoding phase, and the confidence of EOS is significantly higher than that of other non-EOS. Therefore, this may lead to inaccurate length control during the early decoding stage when using EOS confidence
In contrast, \textbf{$\rho$-\texttt{EOS}} provides a more global and aggregated estimate of termination likelihood across masked positions, resulting in more reliable adjustment decisions. 
Moreover, the results on two-stage demonstrates that implicit \texttt{EOS} density, $\rho$, is also applicable as a length control signal for two-stage strategies, further amplifies $\rho$ advantage. 
Overall, these results demonstrate that \textbf{implicit \texttt{EOS} density is a superior control signal compared to confidence}.

\textbf{The convergence trend of $\rho$-\texttt{EOS}.} 
Figure~\ref{fig:visualization convergence rho EOS} visualizes the evolution of the implicit \texttt{EOS} density throughout the denoising process. Across different tasks and initial lengths, we observe that for fixed-length baselines, a short length causes $\rho_\texttt{EOS}$ to converge towards 0, while a long length causes $\rho_\texttt{EOS}$ to converge towards 1. These two indicate insufficient and excessive masked space, respectively. Only when the length is appropriate, it will converge to a length-equilibrium region.
In contrast \textbf{$\rho$-\texttt{EOS}} (red, orange and yellow bar) exhibits a clear convergence pattern: after a small number of denoising steps, the estimated density stabilizes within the length-equilibrium region before final decoding.

%% file: tables/main_results_1.tex
\begin{figure*}[t!]
    \centering
    \begin{minipage}[t]{0.69\textwidth}
        \centering
        \renewcommand{\arraystretch}{1.1}
        \begin{table}[H]
        \caption{\textbf{Main Results on LLaDA-1.5-8B across Four Benchmarks.} We compare the \textbf{$\rho$-\texttt{EOS}} performance against DAEDAL and various baseline configurations. $\boldsymbol{Acc}$ denotes accuracy, $\boldsymbol{E_\text{token}}$ is the average effective tokens (the response length excluding trailing padding), $\boldsymbol{N_\text{token}}$ is the average total tokens, and $\boldsymbol{E_\text{ratio}}$ is the effective token ratio. $\boldsymbol{T_\text{runtime}}$ represents the runtime spent on the evaluation (in seconds). The best configuration for the baseline is highlighted in \textcolor{baseline_orange}{\textbf{orange}}. DAEDAL is highlighted in \textcolor{daedal_blue}{\textbf{blue}}, and \textbf{$\rho$-\texttt{EOS}} is highlighted in \textcolor{rho_EOS_red}{\textbf{red}}. Under the \textbf{$\rho$-\texttt{EOS}} setting, Sym and Asym denote the symmetric and asymmetric lower-upper threshold centered around 0.5. The \textbf{best} results are \textbf{bold}, and the \underline{second-best} results are \underline{underlined}.}
        \label{tab:main_results_1}
        \vspace{-0.1cm}
        \resizebox{0.99\textwidth}{!}{
        \begin{tabular}{llccccccccc}
        \toprule
        \multirow{2}{*}{\textbf{Benchmark}} & \multirow{2}{*}{\textbf{Metric}} & \multicolumn{6}{c}{\textbf{Fixed-Length Denoising (Baseline)}} & \multicolumn{1}{c}{\cellcolor{lightblue}\textbf{DAEDAL}} & {\cellcolor{lightred}\textbf{$\rho$-\texttt{EOS}}} & {\cellcolor{lightred}\textbf{$\rho$-\texttt{EOS}}} \\
        \cmidrule(lr){3-11}
        & & 64 & 128 & 256 & 512 & 1024 & 2048 & \multicolumn{1}{c}{\cellcolor{lightblue}{64}} & \multicolumn{1}{c}{\cellcolor{lightred}{64 Sym}} & \multicolumn{1}{c}{\cellcolor{lightred}{64 Asym}} \\
        \midrule
        \multirow{5}{*}{\textbf{GSM8K}} & $\boldsymbol{Acc}$ & 49.9 & 71.2 & 80.5 & 83.6 & 83.9 & \cellcolor{lightorange}{84.2} & \cellcolor{lightblue}\textbf{85.1} & \cellcolor{lightred}{83.3} & \cellcolor{lightred}\underline{84.5} \\
        & $\boldsymbol{E_\text{token}}$ & 62.4 & 124.8 & 236.7 & 292.6 & 287 & \cellcolor{lightorange}{293.9} & \cellcolor{lightblue}{254.7} & \cellcolor{lightred}{214.5} & \cellcolor{lightred}{231.6} \\
        & $\boldsymbol{N_\text{token}}$ & 64 & 128 & 256 & 512 & 1024 & \cellcolor{lightorange}{2048} & \cellcolor{lightblue}{348.3} & \cellcolor{lightred}{254.3} & \cellcolor{lightred}{362.4} \\
        & $\boldsymbol{E_\text{ratio}}$ & 97.1\% & 97.0\% & 91.2\% & 56.0\% & 27.7\% & \cellcolor{lightorange}{14.4\%} & \cellcolor{lightblue}\underline{72.9\%} & \cellcolor{lightred}\textbf{84.4\%} & \cellcolor{lightred}{63.9\%} \\
        & $\boldsymbol{T_\text{runtime}}$ & 149 & 335 & 844 & 2488 & 8450 & \cellcolor{lightorange}{30875} & \cellcolor{lightblue}{1056} & \cellcolor{lightred}\textbf{721} & \cellcolor{lightred}\underline{990} \\
        \midrule
        \multirow{5}{*}{\textbf{MATH500}} & $\boldsymbol{Acc}$ & 22.8 & 30.2 & 35.2 & 39.2 & \cellcolor{lightorange}\textbf{42.8} & 39.4 & \cellcolor{lightblue}\underline{42.6} & \cellcolor{lightred}{38.2} & \cellcolor{lightred}{41.0} \\
        & $\boldsymbol{E_\text{token}}$ & 62.1 & 124.8 & 246.5 & 428.6 & \cellcolor{lightorange}{583.5} & 713.7 & \cellcolor{lightblue}{489.1} & \cellcolor{lightred}{339.5} & \cellcolor{lightred}{583.4} \\
        & $\boldsymbol{N_\text{token}}$ & 64 & 128 & 256 & 512 & \cellcolor{lightorange}{1024} & 2048 & \cellcolor{lightblue}662.6 & \cellcolor{lightred}{373.1} & \cellcolor{lightred}{711.3} \\
        & $\boldsymbol{E_\text{ratio}}$ & 97.0\% & 97.5\% & 96.3\% & 83.7\% & \cellcolor{lightorange}{57.0\%} & 34.9\% & \cellcolor{lightblue}{73.8\%} & \cellcolor{lightred}\textbf{91.0\%} & \cellcolor{lightred}\underline{82.0\%} \\
        & $\boldsymbol{T_\text{runtime}}$ & 97 & 192 & 444 & 1177 & \cellcolor{lightorange}{3616} & 12610 & \cellcolor{lightblue}{2460} & \cellcolor{lightred}\underline{850} & \cellcolor{lightred}\textbf{637} \\
        \midrule
        \multirow{5}{*}{\textbf{MBPP}} & $\boldsymbol{Acc}$ & 21.2 & 30.4 & 39.2 & 38.8 & \cellcolor{lightorange}\underline{39.4} & 39.4 & \cellcolor{lightblue}{38.2} & \cellcolor{lightred}\textbf{40.8} & \cellcolor{lightred}\textbf{40.8} \\
        & $\boldsymbol{E_\text{token}}$ & 60.9 & 123.7 & 238.2 & 347.3 & \cellcolor{lightorange}{343.5} & 356.2 & \cellcolor{lightblue}{321.8} & \cellcolor{lightred}{336.1} & \cellcolor{lightred}{334.1} \\
        & $\boldsymbol{N_\text{token}}$ & 64 & 128 & 256 & 512 & \cellcolor{lightorange}{1024} & 2048 & \cellcolor{lightblue}{587.1} & \cellcolor{lightred}{535.6} & \cellcolor{lightred}{540.8} \\
        & $\boldsymbol{E_\text{ratio}}$ & 95.1\% & 96.6\% & 93.0\% & 67.8\% & \cellcolor{lightorange}{33.6\%} & 17.4\% & \cellcolor{lightblue}{54.8\%} & \cellcolor{lightred}\textbf{62.8\%} & \cellcolor{lightred}\underline{61.8\%} \\
        & $\boldsymbol{T_\text{runtime}}$ & 156 & 313 & 690 & 1689 & \cellcolor{lightorange}{4688} & 14895 & \cellcolor{lightblue}{4124} & \cellcolor{lightred}\underline{3001} & \cellcolor{lightred}\textbf{2565} \\
        \midrule
        \multirow{5}{*}{\textbf{HUMANEVAL}} & $\boldsymbol{Acc}$ & 17.6 & 20.7 & 38.4 & 45.1 & \cellcolor{lightorange}\textbf{49.4} & 48.8 & \cellcolor{lightblue}{43.9} & \cellcolor{lightred}\underline{44.5} & \cellcolor{lightred}\underline{44.5} \\
        & $\boldsymbol{E_\text{token}}$ & 60.3 & 124.9 & 251.7 & 473.8 & \cellcolor{lightorange}672.1 & 761.6 & \cellcolor{lightblue}{523} & \cellcolor{lightred}{460.8} & \cellcolor{lightred}{686.6} \\
        & $\boldsymbol{N_\text{token}}$ & 64 & 128 & 256 & 512 & \cellcolor{lightorange}{1024} & 2048 & \cellcolor{lightblue}{806.4} & \cellcolor{lightred}{509.7} & \cellcolor{lightred}{809.5} \\
        & $\boldsymbol{E_\text{ratio}}$ & 94.2\% & 97.5\% & 98.3\% & 92.5\% & \cellcolor{lightorange}{65.6\%} & 37.2\% & \cellcolor{lightblue}{64.4\%} & \cellcolor{lightred}\textbf{90.4\%} & \cellcolor{lightred}\underline{84.8\%} \\
        & $\boldsymbol{T_\text{runtime}}$ & 47 & 78 & 162 & 406 & \cellcolor{lightorange}{1212} & 4187 & \cellcolor{lightblue}{743} & \cellcolor{lightred}\textbf{433} & \cellcolor{lightred}\underline{647} \\
        \bottomrule
        \end{tabular}
        }
        \end{table}
    \end{minipage}
    \begin{minipage}[t]{0.305\textwidth}
    \vspace{12pt}
        \centering
        \includegraphics[width=0.88\textwidth]{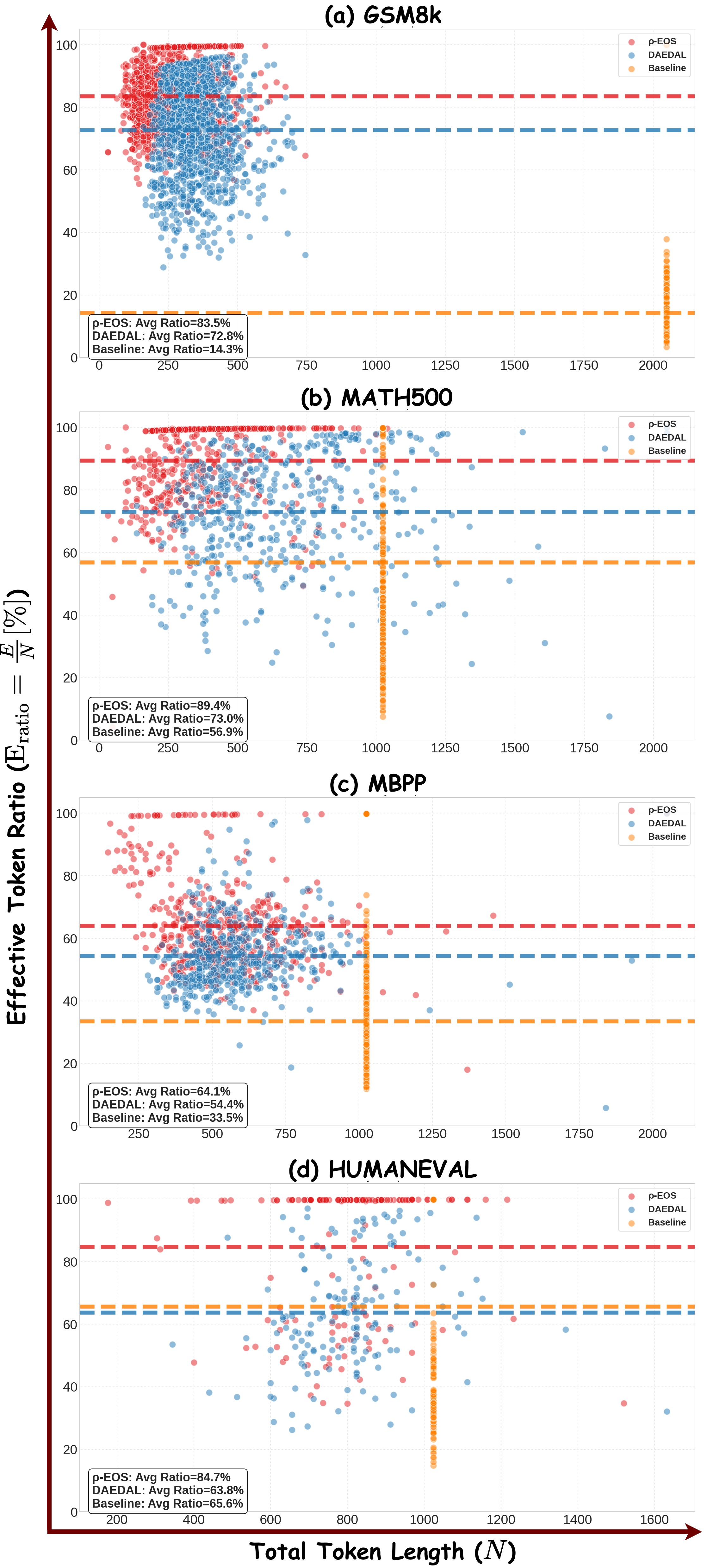}
        \vspace{-0.2cm}
        \caption{$\mathrm{E}_\text{ratio}$ per sample.} 
        \label{fig:LLaDA_1p5_E_ratio}
    \end{minipage}
\vspace{-0.4cm}
\end{figure*}

%% file: tables/var_init_len.tex
\renewcommand{\arraystretch}{1.1}
\begin{table*}[!t]
    \caption{\textbf{Various-Initial Length Results on LLaDA-Instruct-8B and LLaDA-1.5-8B.} We compare \textbf{$\rho$-\texttt{EOS}} at various-initial lengths (128 to 1024) against DAEDAL. The \textbf{best} results are \textbf{bold}, and the \underline{second-best} results are \underline{underlined}.}
    \label{tab:var_init_len}
    \vspace{-0.3cm}
    \begin{center}
    \resizebox{1.0\textwidth}{!}{
    \begin{tabular}{llcc cc cc cc|cc||cc cc cc cc|cc}
        \toprule
        \multirow{3}{*}{\textbf{Benchmark}} & \multirow{3}{*}{\textbf{Metric}} & \multicolumn{10}{c||}{\textbf{LLaDA Various-Init-Length}} & \multicolumn{10}{c}{\textbf{LLaDA1.5 Various-Init-Length}} \\
        \cmidrule(lr){3-22}
        & & \multicolumn{2}{c|}{128} & \multicolumn{2}{c|}{256} & \multicolumn{2}{c|}{512} & \multicolumn{2}{c|}{1024} & \multicolumn{2}{c||}{Average} & \multicolumn{2}{c|}{128} & \multicolumn{2}{c|}{256} & \multicolumn{2}{c|}{512} & \multicolumn{2}{c|}{1024} & \multicolumn{2}{c}{Average} \\
        \cmidrule(lr){3-22}
        & & \cellcolor{lightblue}\textbf{DAEDAL} & \cellcolor{lightred}\textbf{$\rho$-\texttt{EOS}} & \cellcolor{lightblue}\textbf{DAEDAL} & \cellcolor{lightred}\textbf{$\rho$-\texttt{EOS}} & \cellcolor{lightblue}\textbf{DAEDAL} & \cellcolor{lightred}\textbf{$\rho$-\texttt{EOS}} & \cellcolor{lightblue}\textbf{DAEDAL} & \cellcolor{lightred}\textbf{$\rho$-\texttt{EOS}} & \cellcolor{lightblue}\textbf{DAEDAL} & \cellcolor{lightred}\textbf{$\rho$-\texttt{EOS}} & \cellcolor{lightblue}\textbf{DAEDAL} & \cellcolor{lightred}\textbf{$\rho$-\texttt{EOS}} & \cellcolor{lightblue}\textbf{DAEDAL} & \cellcolor{lightred}\textbf{$\rho$-\texttt{EOS}} & \cellcolor{lightblue}\textbf{DAEDAL} & \cellcolor{lightred}\textbf{$\rho$-\texttt{EOS}} & \cellcolor{lightblue}\textbf{DAEDAL} & \cellcolor{lightred}\textbf{$\rho$-\texttt{EOS}} & \cellcolor{lightblue}\textbf{DAEDAL} & \cellcolor{lightred}\textbf{$\rho$-\texttt{EOS}} \\
        \midrule
        \multirow{5}{*}{\textbf{GSM8K}} & $\boldsymbol{Acc}$ & \cellcolor{lightblue}{84.6} & \cellcolor{lightred}{84.6} & \cellcolor{lightblue}{84.0} & \cellcolor{lightred}{83.7} & \cellcolor{lightblue}{85.3} & \cellcolor{lightred}{84.4} & \cellcolor{lightblue}{84.8} & \cellcolor{lightred}{84.8} & 
        \cellcolor{lightblue}\textbf{84.7} & \cellcolor{lightred}\underline{84.4} & 
        \cellcolor{lightblue}{85.1} & \cellcolor{lightred}{85.2} & \cellcolor{lightblue}{85.0} & \cellcolor{lightred}{83.8} & \cellcolor{lightblue}{84.2} & \cellcolor{lightred}{84.3} & \cellcolor{lightblue}{85.9} & \cellcolor{lightred}{84.9} & \cellcolor{lightblue}\textbf{85.1} & \cellcolor{lightred}\underline{84.5} \\
        & $\boldsymbol{E_\text{token}}$ & 
        \cellcolor{lightblue}{246.3} &\cellcolor{lightred}{231.8} & \cellcolor{lightblue}{250.2} &\cellcolor{lightred}{284.7} & \cellcolor{lightblue}{286.8} &\cellcolor{lightred}{288.6} & \cellcolor{lightblue}{281.3} &\cellcolor{lightred}{282.6} & \cellcolor{lightblue}{266.2} &\cellcolor{lightred}{271.9} &
        \cellcolor{lightblue}{254.7} &\cellcolor{lightred}{257.8} & \cellcolor{lightblue}{256.9} &\cellcolor{lightred}{261.0} & \cellcolor{lightblue}{275.7} &\cellcolor{lightred}{293.7} & \cellcolor{lightblue}{284.0} &\cellcolor{lightred}{292.5} & \cellcolor{lightblue}{267.8} &\cellcolor{lightred}{276.3} \\
        & $\boldsymbol{N_\text{token}}$ & 
        \cellcolor{lightblue}{330.8} &\cellcolor{lightred}{313.0} & \cellcolor{lightblue}{341.1} &\cellcolor{lightred}{336.4} & \cellcolor{lightblue}{530.0} &\cellcolor{lightred}{438.6} & \cellcolor{lightblue}{1040.0} &\cellcolor{lightred}{666.8} & \cellcolor{lightblue}{560.5} &\cellcolor{lightred}{438.7} &
        \cellcolor{lightblue}{349.3} &\cellcolor{lightred}{370.5} & \cellcolor{lightblue}{356.4} &\cellcolor{lightred}{321.0} & \cellcolor{lightblue}{531.4} &\cellcolor{lightred}{474.2} & \cellcolor{lightblue}{1040.0} &\cellcolor{lightred}{398.0} & \cellcolor{lightblue}{569.3} &\cellcolor{lightred}{390.9} \\
        & $\boldsymbol{E_\text{ratio}}$ &
        \cellcolor{lightblue}{74.5\%}&\cellcolor{lightred}{74.1\%}&
        \cellcolor{lightblue}{73.4\%}&\cellcolor{lightred}{84.6\%}& 
        \cellcolor{lightblue}{50.9\%}&\cellcolor{lightred}{65.8\%}& \cellcolor{lightblue}{27.0\%}&\cellcolor{lightred}{42.4\%}&
        \cellcolor{lightblue}\underline{56.5\%}&
        \cellcolor{lightred}\textbf{66.7\%}&
        \cellcolor{lightblue}{72.9\%}&\cellcolor{lightred}{69.6\%}&
        \cellcolor{lightblue}{72.1\%}&\cellcolor{lightred}{81.3\%}&
        \cellcolor{lightblue}{51.9\%}&\cellcolor{lightred}{61.9\%}&
        \cellcolor{lightblue}{27.3\%}&\cellcolor{lightred}{73.5\%}&
        \cellcolor{lightblue}\underline{56.1\%}&\cellcolor{lightred}\textbf{71.6\%}\\
        & $\boldsymbol{T_\text{runtime}}$ &
        \cellcolor{lightblue}{1035} & \cellcolor{lightred}{835} & \cellcolor{lightblue}{1014} & \cellcolor{lightred}{834} & \cellcolor{lightblue}{1090} & \cellcolor{lightred}{1044} & \cellcolor{lightblue}{5656} & \cellcolor{lightred}{1809} & \cellcolor{lightblue}\underline{2199.0} & 
        \cellcolor{lightred}\textbf{1131.0} & 
        \cellcolor{lightblue}{1093} & \cellcolor{lightred}{990} & \cellcolor{lightblue}{1008} & \cellcolor{lightred}{638} & \cellcolor{lightblue}{1223} & \cellcolor{lightred}{1122} & \cellcolor{lightblue}{5476} & \cellcolor{lightred}{1281} & \cellcolor{lightblue}\underline{2200.0} & \cellcolor{lightred}\textbf{1007.8} \\
        \midrule
        \multirow{5}{*}{\textbf{MATH500}} & $\boldsymbol{Acc}$ & \cellcolor{lightblue}{42.6} & \cellcolor{lightred}{40.0} & \cellcolor{lightblue}{43.2} & \cellcolor{lightred}{40.4} & \cellcolor{lightblue}{41.8} & \cellcolor{lightred}{42.8} & \cellcolor{lightblue}{41.6} & \cellcolor{lightred}{39.8} &
        \cellcolor{lightblue}\textbf{42.3} & \cellcolor{lightred}\underline{40.8} &
        \cellcolor{lightblue}{42.6} & \cellcolor{lightred}{40.6} & \cellcolor{lightblue}{43.0} & \cellcolor{lightred}{41.2} & \cellcolor{lightblue}{41.4} & \cellcolor{lightred}{42.0} & \cellcolor{lightblue}{40.0} & \cellcolor{lightred}{40.6} & \cellcolor{lightblue}\textbf{41.8} & \cellcolor{lightred}\underline{41.1} \\
        & $\boldsymbol{E_\text{token}}$ & 
        \cellcolor{lightblue}{464.6} &\cellcolor{lightred}{410.8} & \cellcolor{lightblue}{465.7} &\cellcolor{lightred}{492.2} & \cellcolor{lightblue}{479.0} &\cellcolor{lightred}{556.6} & \cellcolor{lightblue}{541.4} &\cellcolor{lightred}{588.4} &
        \cellcolor{lightblue}{487.7} &\cellcolor{lightred}{512.0} &
        \cellcolor{lightblue}{489.1} &\cellcolor{lightred}{528.1} & \cellcolor{lightblue}{490.0} &\cellcolor{lightred}{559.2} & \cellcolor{lightblue}{499.4} &\cellcolor{lightred}{586.7} & \cellcolor{lightblue}{534.3} &\cellcolor{lightred}{602.3} & \cellcolor{lightblue}{503.2} &\cellcolor{lightred}{569.1} \\
        & $\boldsymbol{N_\text{token}}$ & 
        \cellcolor{lightblue}{618.4} &\cellcolor{lightred}{480.8} & \cellcolor{lightblue}{621.0} &\cellcolor{lightred}{549.2} & \cellcolor{lightblue}{686.8} &\cellcolor{lightred}{642.0} & \cellcolor{lightblue}{1058.8}&\cellcolor{lightred}{874.1} & 
        \cellcolor{lightblue}{746.3} &\cellcolor{lightred}{636.5} &
        \cellcolor{lightblue}{662.6} &\cellcolor{lightred}{624.1} & \cellcolor{lightblue}{664.2} &\cellcolor{lightred}{664.3} & \cellcolor{lightblue}{717.7} &\cellcolor{lightred}{692.9} & \cellcolor{lightblue}{1064.9}&\cellcolor{lightred}{792.1} & \cellcolor{lightblue}{777.4}&\cellcolor{lightred}{693.4} \\
        & $\boldsymbol{E_\text{ratio}}$ & 
        \cellcolor{lightblue}{75.1\%}&\cellcolor{lightred}{89.2\%}& \cellcolor{lightblue}{75.0\%}&\cellcolor{lightred}{89.5\%}& 
        \cellcolor{lightblue}{69.7\%}&\cellcolor{lightred}{86.7\%}& \cellcolor{lightblue}{51.1\%}&\cellcolor{lightred}{67.3\%}&
        \cellcolor{lightblue}\underline{67.7\%}&
        \cellcolor{lightred}\textbf{83.2\%}&
        \cellcolor{lightblue}{73.8\%}&\cellcolor{lightred}{84.6\%}& \cellcolor{lightblue}{73.8\%}&\cellcolor{lightred}{84.2\%}& \cellcolor{lightblue}{69.6\%}&\cellcolor{lightred}{84.7\%}& \cellcolor{lightblue}{50.2\%}&\cellcolor{lightred}{76.0\%}& \cellcolor{lightblue}\underline{66.9\%}&
        \cellcolor{lightred}\textbf{82.4\%}\\
        & $\boldsymbol{T_\text{runtime}}$ & 
        \cellcolor{lightblue}{2199} & \cellcolor{lightred}{1062} & \cellcolor{lightblue}{2175} & \cellcolor{lightred}{1543} & \cellcolor{lightblue}{2096} & \cellcolor{lightred}{1980} & \cellcolor{lightblue}{6543} & \cellcolor{lightred}{1938} & 
        \cellcolor{lightblue}\underline{3253.3} & \cellcolor{lightred}\textbf{1630.8} &
        \cellcolor{lightblue}{2447} & \cellcolor{lightred}{1805} & \cellcolor{lightblue}{2423} & \cellcolor{lightred}{1912} & \cellcolor{lightblue}{2345} & \cellcolor{lightred}{2074} & \cellcolor{lightblue}{2315} & \cellcolor{lightred}{2199} & \cellcolor{lightblue}\underline{2382.5} & \cellcolor{lightred}\textbf{1997.5}\\
        \midrule
        \multirow{5}{*}{\textbf{MBPP}} 
        & $\boldsymbol{Acc}$ & 
        \cellcolor{lightblue}{39.4} & \cellcolor{lightred}{39.0} & \cellcolor{lightblue}{39.4} & \cellcolor{lightred}{40.0} & 
        \cellcolor{lightblue}{38.8} & \cellcolor{lightred}{39.4} & \cellcolor{lightblue}{37.8} & \cellcolor{lightred}{38.6} & 
        \cellcolor{lightblue}\underline{38.8} & 
        \cellcolor{lightred}\textbf{39.3} &
        \cellcolor{lightblue}{38.2} & \cellcolor{lightred}{41.8} & \cellcolor{lightblue}{38.2} & \cellcolor{lightred}{41.4} & \cellcolor{lightblue}{37.8} & \cellcolor{lightred}{39.0} & \cellcolor{lightblue}{40.0} & \cellcolor{lightred}{40.4} & \cellcolor{lightblue}\underline{38.6} & \cellcolor{lightred}\textbf{40.7} \\
        & $\boldsymbol{E_\text{token}}$ & 
        \cellcolor{lightblue}{308.4} &\cellcolor{lightred}{341.6} & \cellcolor{lightblue}{308.4} &\cellcolor{lightred}{344.0} & \cellcolor{lightblue}{313.0} &\cellcolor{lightred}{342.5} & \cellcolor{lightblue}{324.7} &\cellcolor{lightred}{345.0} &
        \cellcolor{lightblue}{313.6} &\cellcolor{lightred}{343.3} &
        \cellcolor{lightblue}{321.8} &\cellcolor{lightred}{345.1} & \cellcolor{lightblue}{321.8} &\cellcolor{lightred}{350.2} & \cellcolor{lightblue}{325.4} &\cellcolor{lightred}{362.1} & \cellcolor{lightblue}{331.5} &\cellcolor{lightred}{350.7} & \cellcolor{lightblue}{325.1} &\cellcolor{lightred}{352.0}\\
        & $\boldsymbol{N_\text{token}}$ & 
        \cellcolor{lightblue}{558.3} &\cellcolor{lightred}{605.1} & \cellcolor{lightblue}{558.3} &\cellcolor{lightred}{611.6} & \cellcolor{lightblue}{604.6} &\cellcolor{lightred}{638.3} & \cellcolor{lightblue}{1046.7}&\cellcolor{lightred}{773.5} &
        \cellcolor{lightblue}{692.0} &\cellcolor{lightred}{657.1} &
        \cellcolor{lightblue}{587.1} &\cellcolor{lightred}{556.5} & \cellcolor{lightblue}{587.1} &\cellcolor{lightred}{497.7} & \cellcolor{lightblue}{621.4} &\cellcolor{lightred}{662.8}& \cellcolor{lightblue}{1046.8}&\cellcolor{lightred}{943.6} & \cellcolor{lightblue}{710.6}&\cellcolor{lightred}{665.2} \\
        & $\boldsymbol{E_\text{ratio}}$ & 
        \cellcolor{lightblue}{55.2\%}&\cellcolor{lightred}{56.5\%}& \cellcolor{lightblue}{55.2\%}&\cellcolor{lightred}{56.3\%}& \cellcolor{lightblue}{51.8\%}&\cellcolor{lightred}{53.6\%}& \cellcolor{lightblue}{31.0\%}&\cellcolor{lightred}{44.6\%}& 
        \cellcolor{lightblue}\underline{48.3\%}&
        \cellcolor{lightred}\textbf{52.8\%}&
        \cellcolor{lightblue}{54.8\%}&\cellcolor{lightred}{62.0\%}& \cellcolor{lightblue}{54.8\%}&\cellcolor{lightred}{70.4\%}& \cellcolor{lightblue}{52.4\%}&\cellcolor{lightred}{54.6\%}& \cellcolor{lightblue}{31.7\%}&\cellcolor{lightred}{37.2\%}& \cellcolor{lightblue}\underline{48.4\%}&
        \cellcolor{lightred}\textbf{56.1\%}\\
        & $\boldsymbol{T_\text{runtime}}$ 
        & \cellcolor{lightblue}{3518} & \cellcolor{lightred}{2836} & \cellcolor{lightblue}{3462} & \cellcolor{lightred}{2169} & \cellcolor{lightblue}{3303} & \cellcolor{lightred}{2389} & \cellcolor{lightblue}{5674} & \cellcolor{lightred}{2558} &
        \cellcolor{lightblue}\underline{3989.3} & \cellcolor{lightred}\textbf{2396.0} &
        \cellcolor{lightblue}{4102} & \cellcolor{lightred}{2824} & \cellcolor{lightblue}{4060} & \cellcolor{lightred}{1853} & \cellcolor{lightblue}{3907} & \cellcolor{lightred}{2585} & \cellcolor{lightblue}{4471} & \cellcolor{lightred}{2501} & \cellcolor{lightblue}\underline{4135.0} & \cellcolor{lightred}\textbf{2440.8}\\
        \midrule
        \multirow{5}{*}{\textbf{HUMANEVAL}} & $\boldsymbol{Acc}$ & \cellcolor{lightblue}{48.2} & \cellcolor{lightred}{48.8} & \cellcolor{lightblue}{48.2} & \cellcolor{lightred}{47.0} & \cellcolor{lightblue}{48.2} & \cellcolor{lightred}{45.1} & \cellcolor{lightblue}{47.0} & \cellcolor{lightred}{47.6} & 
        \cellcolor{lightblue}\textbf{47.9} &
        \cellcolor{lightred}\underline{47.1} &
        \cellcolor{lightblue}{43.9} & \cellcolor{lightred}{47.6} & \cellcolor{lightblue}{43.9} & \cellcolor{lightred}{47.0} & \cellcolor{lightblue}{43.9} & \cellcolor{lightred}{48.2} & \cellcolor{lightblue}{43.9} & \cellcolor{lightred}{49.4} & \cellcolor{lightblue}\underline{43.9} & 
        \cellcolor{lightred}\textbf{48.1}\\
        & $\boldsymbol{E_\text{token}}$ & 
        \cellcolor{lightblue}{497.5} &\cellcolor{lightred}{618.0} & \cellcolor{lightblue}{497.5} &\cellcolor{lightred}{613.0} & \cellcolor{lightblue}{497.3} &\cellcolor{lightred}{613.7} & \cellcolor{lightblue}{540.0} &\cellcolor{lightred}{638.8} &
        \cellcolor{lightblue}{508.1} &\cellcolor{lightred}{620.9} &
        \cellcolor{lightblue}{519.2} &\cellcolor{lightred}{634.7} & \cellcolor{lightblue}{519.2} &\cellcolor{lightred}{654.5} & \cellcolor{lightblue}{518.8} &\cellcolor{lightred}{623.6} & \cellcolor{lightblue}{528.4} &\cellcolor{lightred}{677.6} & \cellcolor{lightblue}{521.4} &\cellcolor{lightred}{647.6} \\
        & $\boldsymbol{N_\text{token}}$ & 
        \cellcolor{lightblue}{769.5} &\cellcolor{lightred}{693.5} & \cellcolor{lightblue}{769.5} &\cellcolor{lightred}{698.7} & \cellcolor{lightblue}{772.4} &\cellcolor{lightred}{717.8} & \cellcolor{lightblue}{1046.9}&\cellcolor{lightred}{927.9} & \cellcolor{lightblue}{839.6}&\cellcolor{lightred}{759.5} &
        \cellcolor{lightblue}{806.4} &\cellcolor{lightred}{734.9} & \cellcolor{lightblue}{806.4} &\cellcolor{lightred}{743.0} & \cellcolor{lightblue}{808.4} &\cellcolor{lightred}{692.2} & \cellcolor{lightblue}{1049.4}&\cellcolor{lightred}{951.5}& \cellcolor{lightblue}{867.7}&\cellcolor{lightred}{780.4}\\
        & $\boldsymbol{E_\text{ratio}}$ & 
        \cellcolor{lightblue}{64.7\%}&\cellcolor{lightred}{89.1\%}& \cellcolor{lightblue}{64.7\%}&\cellcolor{lightred}{87.7\%}& \cellcolor{lightblue}{64.4\%}&\cellcolor{lightred}{85.5\%}& \cellcolor{lightblue}{51.6\%}&\cellcolor{lightred}{68.8\%}& \cellcolor{lightblue}\underline{61.4\%}&
        \cellcolor{lightred}\textbf{82.8\%}&
        \cellcolor{lightblue}{64.4\%}&\cellcolor{lightred}{86.4\%}& \cellcolor{lightblue}{64.4\%}&\cellcolor{lightred}{88.1\%}& \cellcolor{lightblue}{64.2\%}&\cellcolor{lightred}{90.1\%}& \cellcolor{lightblue}{50.4\%}&\cellcolor{lightred}{71.2\%}& \cellcolor{lightblue}\underline{60.9\%}&
        \cellcolor{lightred}\textbf{84.0\%}\\
        & $\boldsymbol{T_\text{runtime}}$ 
        & \cellcolor{lightblue}{746} & \cellcolor{lightred}{528} & \cellcolor{lightblue}{789} & \cellcolor{lightred}{542} & \cellcolor{lightblue}{725} & \cellcolor{lightred}{548} & \cellcolor{lightblue}{1644} & \cellcolor{lightred}{618} &
        \cellcolor{lightblue}\underline{976.0} & 
        \cellcolor{lightred}\textbf{559.0} &
        \cellcolor{lightblue}{786} & \cellcolor{lightred}{590} & \cellcolor{lightblue}{730} & \cellcolor{lightred}{593} & \cellcolor{lightblue}{705} & \cellcolor{lightred}{623} & \cellcolor{lightblue}{798} & \cellcolor{lightred}{606} & \cellcolor{lightblue}\underline{754.8} &
        \cellcolor{lightred}\textbf{603.0} \\
        \bottomrule
    \end{tabular}
    }
    \end{center}
    \vspace{-0.5cm}
\end{table*}

%% file: tables/ablation_rho.tex
\renewcommand{\arraystretch}{1}
\begin{table}[!t]
\centering
    \caption{\textbf{Ablation on $\rho_\text{high}$ and $\rho_\text{low}$.} Ablations are conducted on GSM8k with LLaDA-Instruct-8B, and $L_\text{init}=64$. The threshold of density ($\rho$) ranging from 0 to 1. We highlight our best-performance Asym setting ([$\rho_\text{low}=0.4,\,\rho_\text{high}=0.8$]) in \textcolor{rho_EOS_red}{\textbf{red}}.}
    \label{tab:ablation_rho}
    \vspace{-0.2cm}
    \resizebox{0.48\textwidth}{!}{
        \begin{tabular}{l *{4}{w{c}{0.1\textwidth}}}
            \toprule
            \multirow{2}{*}{\textbf{Metric}} & \multicolumn{4}{c}{\textbf{Density Threshold \, [$\rho_\text{low},\,\rho_\text{high}$]}} \\
            \cmidrule(lr){2-5}
            & [0.2, 0.8] & [0.3, 0.7] & [0.4, 0.6] & \cellcolor{lightred}[0.4, 0.8] \\
            \midrule
            $\boldsymbol{Acc}$ & 81.8 & 81.7 & 81.7 & \cellcolor{lightred}84.2 \\
            $\boldsymbol{E_\text{token}}$ & 198.4 & 210.5 & 218.4 & \cellcolor{lightred}252.5 \\
            $\boldsymbol{N_\text{token}}$ & 231.8 & 251.2 & 267.5 & \cellcolor{lightred}361.0 \\
            $\boldsymbol{E_\text{ratio}}$ & 85.6\% & 83.8\% & 81.6\% & \cellcolor{lightred}70.0\% \\
            $\boldsymbol{T_\text{runtime}}$ & 645 & 722 & 845 & \cellcolor{lightred}823 \\
            \bottomrule
            \end{tabular}
    }
    \vspace{-0.6cm}
\end{table}

%% file: tables/ablation_E_factor.tex
\renewcommand{\arraystretch}{1}
\begin{table}[!t]
\centering
    \caption{\textbf{Ablation on Expansion Factor Function.} Ablations are conducted on GSM8K with LLaDA-Instruct-8B, $L_\text{init}=64$ and [$\rho_\text{low}=0.4,\,\rho_\text{high}=0.6$]. $\boldsymbol{Step}$ is the average adjustment steps.}
    \label{tab:ablation_rho}
    \vspace{-0.1cm}
    \resizebox{0.372\textwidth}{!}{
        \begin{tabular}{l cccc}
            \toprule
            \multirow{2}{*}{\textbf{Metric}} & \multicolumn{4}{c}{\textbf{Expansion Factor Function} $\mathrm{E}_\text{factor}(\cdot)$} \\
            \cmidrule(lr){2-5}
             & Baseline$_{256}$ & $\mathrm{E}_\text{const}(\cdot)$ & $\mathrm{E}_\text{linear}(\cdot)$ 
             & $\mathrm{E}_\text{exp}(\cdot)$ \\
            \cmidrule(lr){1-5}
            $\boldsymbol{Acc}$ & 77.8& 79.8 & 80.9 & 81.7 \\
            $\boldsymbol{E_\text{token}}$ & 233.7& 183.4 & 208.0 & 218.4 \\
            $\boldsymbol{N_\text{token}}$ & 256& 213.3 & 251.0 & 267.5\\
            $\boldsymbol{E_\text{ratio}}$ & 91.3\%& 86.0\% & 82.9\% & 81.6\%\\
            $\boldsymbol{T_\text{runtime}}$ & 860 & 589 & 757 & 845\\
            $\boldsymbol{Step}$ & - & 61.0 & 54.8 & 52.4\\
            \bottomrule
            \end{tabular}
    }
    \vspace{-0.7cm}
\end{table}

%% file: tables/rho_vs_confidence.tex
\begin{figure*}[t!]
    \centering
    \begin{minipage}[t]{0.59\textwidth}
        \centering
        \renewcommand{\arraystretch}{1.1}
        \begin{table}[H]
        \centering
            \caption{\textbf{Comparison on \texttt{EOS} Density ($\rho$) and Confidence for Length Control and Single Stage v.s. Two Stage.} Ablations are conducted on GSM8K and MBPP with LLaDA-Instruct-8B, $L_\text{init}=64$.}
            \label{tab:rho_vs_confidence}
            \vspace{-0.2cm}
            \resizebox{0.99\textwidth}{!}{
            \begin{tabular}{lcc|cc||cc|cc}
                \toprule
                \multirow{3}{*}{\textbf{Metric}} & \multicolumn{4}{c}{\textbf{GSM8k}} & \multicolumn{4}{c}{\textbf{MBPP}} \\
                \cmidrule(lr){2-5} \cmidrule(lr){6-9}
                & \multicolumn{2}{c}{\textbf{Single Stage}} & \multicolumn{2}{c}{\textbf{Two Stage}} & \multicolumn{2}{c}{\textbf{Single Stage}} & \multicolumn{2}{c}{\textbf{Two Stage}} \\
                \cmidrule(lr){2-3} \cmidrule(lr){4-5} \cmidrule(lr){6-7} \cmidrule(lr){8-9}
                & Confidence & Density ($\rho$)
                & Confidence & Density ($\rho$) 
                & Confidence & Density ($\rho$) 
                & Confidence & Density ($\rho$) \\
                \cmidrule(lr){2-3} \cmidrule(lr){4-5} \cmidrule(lr){6-7} \cmidrule(lr){8-9}
                $\boldsymbol{Acc}$ 
                & 73.7 & \textbf{84.2} 
                & \textbf{84.6} & 84.4
                & 40.0 & \textbf{40.6}
                & 38.2 & \textbf{40.4} \\
                $\boldsymbol{E_\text{token}}$ 
                & 224.3 & 252.2
                & 246.0 & 219.9
                & 340.7 & 335.9
                & 321.8 & 335.3 \\
                $\boldsymbol{N_\text{token}}$ 
                & 256.6 & 361.0 
                & 331.0 & 414.9
                & 546.5 & 594.4 
                & 587.1 & 602.3 \\
                $\boldsymbol{E_\text{ratio}}$ 
                & 87.4\% & 70.0\% 
                & 74.5\% & 53.0\%
                & 62.3\% & 55.5\%
                & 54.8\% & 55.7\% \\
                $\boldsymbol{T_\text{runtime}}$ 
                & 888 & 823 
                & 1090 & 1049
                & 2560 & 2030
                & 4124 & 2030 \\
                \bottomrule
                \end{tabular}
            }
            \vspace{-0.6cm}
        \end{table}
    \end{minipage}
    \begin{minipage}[t]{0.39\textwidth}
    \vspace{12pt}
        \centering
        \includegraphics[width=0.95\textwidth]{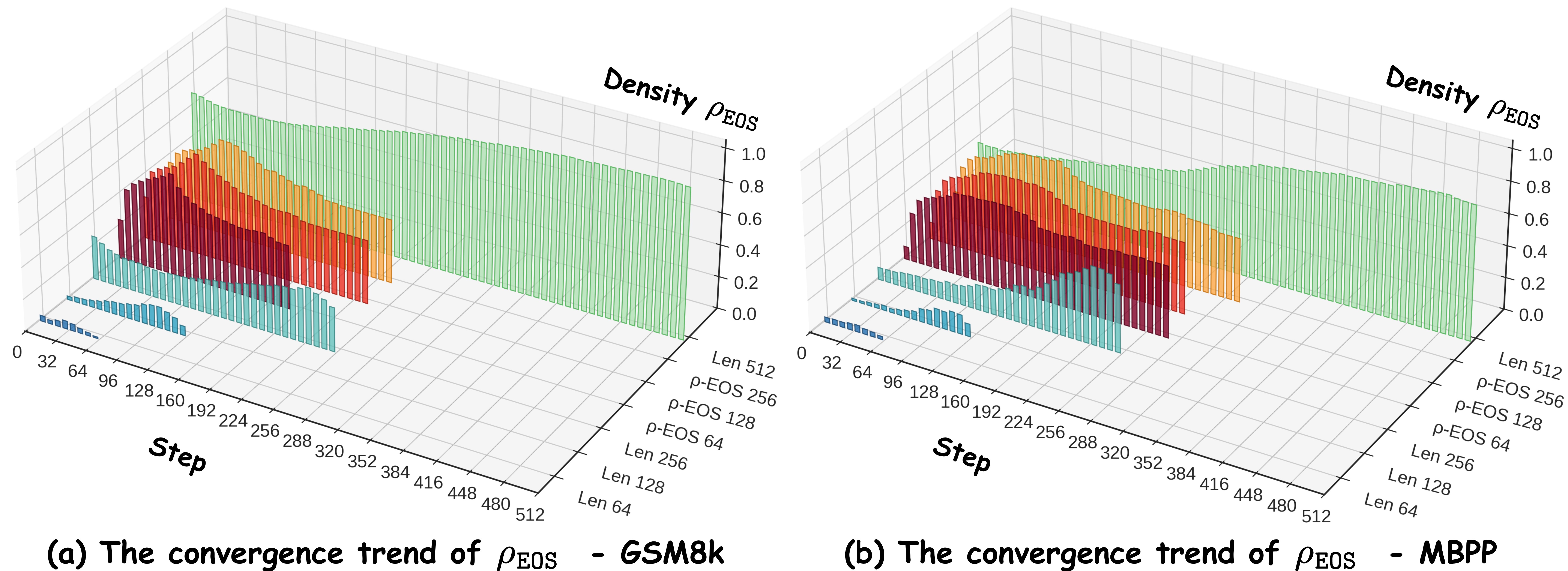} 
        \caption{The convergence trend of $\rho_\texttt{EOS}$ during denoising: baseline with different fixed-length and $\rho$-\texttt{EOS} with different initial lengths.} 
        \label{fig:visualization convergence rho EOS}
    \end{minipage}
\vspace{-0.4cm}
\end{figure*}

%% file: sections/conclusion.tex
\section{Conclusion}
\label{sec:conclusion}

In this work, we propose \textbf{$\rho$-\texttt{EOS}}, a training-free, single-stage bidirectional variable-length denoising strategy for masked diffusion large language models. By leveraging the implicit \texttt{EOS} density signals during denoising, \textbf{$\rho$-\texttt{EOS}} continuously assesses length sufficiency and enables both expansion and contraction within a unified inference loop, eliminating the need for multi-stage.
Extensive experiments on mathematical reasoning and code generation benchmarks show that \textbf{$\rho$-\texttt{EOS}} achieves performance comparable to, and in several cases surpassing, fixed-length baselines and the two-stage strategies, while substantially improving computational efficiency, yielding higher token utilization and reduced inference latency.
Beyond that, our analysis reveals that masked dLLMs inherently possess a strong internal estimate of the appropriate response length for a given task. The implicit \texttt{EOS} density provides a reliable signal for exposing and exploiting this latent capability, allowing fixed-length denoising to be seamlessly transformed into a flexible and adaptive generation process.